\pdfoutput=1
\documentclass[11pt]{article}

\usepackage[final]{coling}

\usepackage{times}
\usepackage{latexsym}
\usepackage{CJKutf8}
\usepackage{graphicx}
\usepackage[T1]{fontenc}

\usepackage[utf8]{inputenc}
\usepackage{booktabs}
\usepackage{microtype}

\usepackage{inconsolata}

\usepackage{graphicx}
\usepackage{amsmath}
\usepackage{fontawesome5}
\usepackage{simpleicons}
\pdfmapfile{+simpleicons.map}
\title{Tangut Word Segmentation under Extreme Resource Scarcity: Integrating Traditional Lexicons and Unlabeled Text}

\author{
  \textbf{Lifan Deng\textsuperscript{1,2}},
  \textbf{Yongwei Zhang\textsuperscript{1,3}\thanks{Corresponding author.}},
  \textbf{Sen Sun\textsuperscript{1,3}},
  \textbf{Bojun Sun\textsuperscript{4}},
  \textbf{Jingsong Yu\textsuperscript{5}}
\\
  \textsuperscript{1}Key Laboratory of Linguistics, \\Chinese Academy of Social Sciences (University of Chinese Academy of Social Sciences), Beijing\\
  \textsuperscript{2}Rixin College, Tsinghua University \\
  \textsuperscript{3}Institute of Linguistics, Chinese Academy of Social Sciences\\
  \textsuperscript{4}Institute of Ethnology and Anthropology, Chinese Academy of Social Sciences\\
  \textsuperscript{5}School of Software and Microelectronics, Peking University\\
  \texttt{zhangyw@cass.org.cn}
}

\begin{document}
\maketitle
\begin{abstract}
Tangut is an extinct language whose script does not explicitly mark word boundaries. We present the first systematic study of Tangut word segmentation using 2,750 expert-annotated segments\footnote{The segments here are not sentences in the linguistic sense; for details, see Appendix~\ref{sec:appendixA1}.} (31,893 tokens), traditional lexicons, and unlabeled text. Our framework combines a reliability-calibrated lexicon-lattice representation, explicit distributional statistics, and a lightweight character encoder
pretrained with MLM. In within-source five-fold cross-validation, the model integrating TangutEncoder, CRF, and external features obtains the numerically highest main-system mean F$_1$ of 0.911 and substantially improves recall beyond the labeled training vocabulary. We further
evaluate document-level transfer on 479 segments (4081 tokens) from five works absent from the annotated training corpus. You can access our project at \url{https://github.com/jiangli-va/TangutSeg}.
\end{abstract}


\section{Introduction}

Created in the 1030s under Li Yuanhao, the Tangut script was used to record the Tangut language and served as the official script of the Xixia. Although Tangut later became extinct, its surviving texts remain essential sources for studying the language, history, and culture of this dynasty.
\begin{figure}[h]
    \centering
    \includegraphics[width=0.9\linewidth]{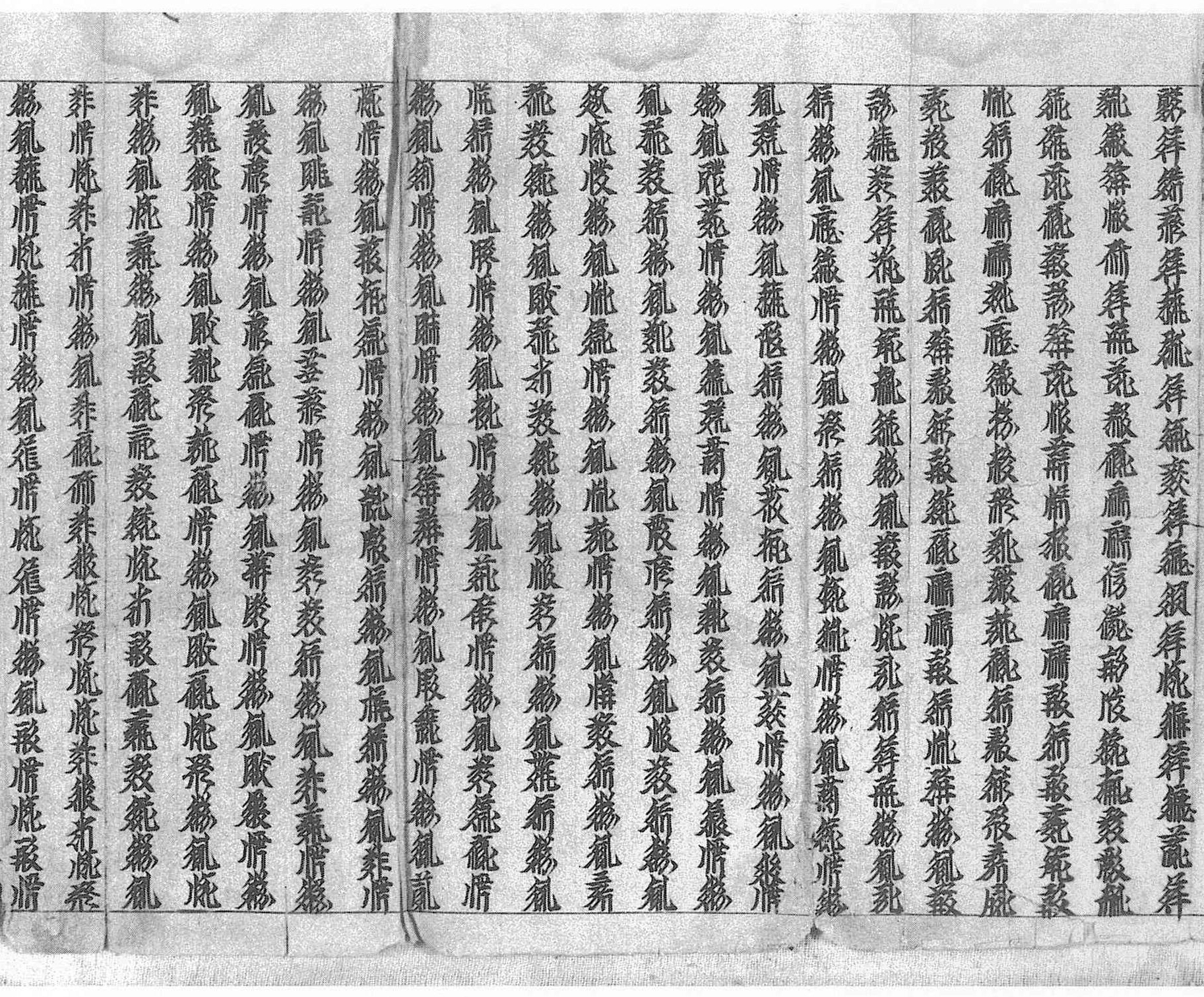}
    \caption{Surviving fragment of \textit{Mahāratnakūṭa-sūtra}, Scroll 68.}
    \label{fig:1}
\end{figure}

Tangut information processing has progressed from character encoding and digital fonts \citep{nakajima-etal-1996-tangut,west-etal-2014-tangut} to optical character recognition \citep{liu-2012-tangut-document-recognition,zhang-han-2017-tangut-recognition,ma-etal-2022-tangut-recognition} and machine translation \citep{zheng-yu-2025-tangut-translation,zheng-etal-2025-tangut-ocr-mt}. However, automatic word segmentation remains largely unexplored. Because Tangut does not explicitly mark word boundaries, OCR and transcription produce unsegmented character sequences. Identifying words is therefore a prerequisite for lexical retrieval, frequency analysis, POS tagging, alignment, and translation.

Tangut word segmentation is constrained by a high annotation barrier. With no native speakers, word boundaries require specialists to jointly consider context, phonological reconstruction, translations, and traditional dictionaries. Our expert-annotated corpus contains 2,750 textual segments and 31,893 word tokens from a Buddhist scripture and the secular encyclopaedic work \textit{Leilin}. Although limited to two works, \textit{Leilin} covers diverse narratives and lexical domains, providing substantial variation for within-source evaluation.

We address these challenges by integrating three sources of knowledge. Expert annotation provides BIES supervision; an aggregated lexicon-lattice representation preserves overlapping dictionary candidates and calibrates their reliability; unlabeled text is exploited through explicit corpus statistics, static character embeddings, and masked-language-model (MLM) pretraining. The resulting lightweight TangutEncoder learns contextual character representations without assuming word boundaries in advance. Under segment-level five-fold cross-validation, the model integrating TangutEncoder, CRF, and external features obtains the numerically highest main-system mean F$_1$ of 0.911 and substantially improves recall beyond the labeled training vocabulary.

Our contributions are threefold:

\begin{itemize}
    \item \textbf{Task and evaluation.} We present the first systematic study of Tangut word segmentation, combining controlled within-source comparisons with independent cross-document evaluation on five additional works.

  \item \textbf{Reliability-aware lexicon integration.} We preserve overlapping dictionary candidates in a lexicon lattice and calibrate their reliability using out-of-fold estimates to prevent boundary leakage.

    \item \textbf{Learning from unlabeled Tangut text.} We compare explicit corpus statistics, static character representations, and contextual MLM pretraining, and develop a lightweight TangutEncoder. 
\end{itemize}

Additionally, we evaluate document-level transfer on an independent test set covering five previously unseen works. These contributions provide a word-level foundation for Tangut text retrieval, POS tagging, alignment, machine translation, and related digital research.

\section{Related Work}
\textbf{Word Segmentation without Explicit Delimiters.}
Segmentation for scripts without word delimiters is commonly formulated as sequence prediction. Chinese work established BIES-style position tagging and CRF segmentation \citep{xue-2003-chinese,peng-etal-2004-chinese}; related formulations use syllable-level CRFs for Tibetan \citep{liu-etal-2011-tibetan} and longest matching for Myanmar \citep{htay-murthy-2008-myanmar}. Neural models subsequently learned candidate and segmentation-history representations \citep{cai-zhao-2016-neural}, convolutional character--word features \citep{wang-xu-2017-convolutional}, and optimized BiLSTM taggers \citep{ma-etal-2018-state}. Yet OOV words remain difficult, and segmentation accuracy depends strongly on orthographic cues and resource availability \citep{shao-etal-2018-universal,brown-2024-improved}. These studies motivate sequence labeling for Tangut but do not address its extreme resource scarcity.

\textbf{Lexicon-Enhanced Segmentation.}
Lexicons have been encoded as CRF features \citep{peng-etal-2004-chinese}, converted into partial annotations for domain adaptation \citep{liu-etal-2014-domain}, and combined with limited manual segmentation for language documentation \citep{okabe-etal-2022-weakly}. Neural segmenters incorporate lexical information through BiLSTMs \citep{Zhang_Liu_Fu_2018}, attention over overlapping candidates \citep{higashiyama-etal-2019-incorporating}, auxiliary BIES prediction from unlabeled text \citep{higashiyama-etal-2020-auxiliary}, or character--word graphs \citep{huang-etal-2021-lexicon-based}. These methods preserve increasingly rich candidate information, but rarely calibrate individual entries whose units may conflict with corpus boundaries.

\textbf{Low-Resource and Historical-Language Segmentation.}
Low-resource NLP lacks not only annotation but often raw text, tools, pretrained models, and sustained support \citep{nigatu-etal-2024-zenos}, limitations pronounced for Sino-Tibetan languages \citep{liu-best-2025-survey}. Dictionary-based weak supervision can partly compensate for scarce segmentation annotation \citep{okabe-etal-2022-weakly}. For historical languages, prior work compares rule-based, lexical, statistical, and learned segmentation for Akkadian \citep{homburg-chiarcos-2016-word}, or uses phonological priors for ancient scripts \citep{luo-etal-2021-deciphering}. Sanskrit research combines expert lexical resources, character models, and latent-candidate Transformers \citep{krishna-etal-2017-dataset,hellwig-nehrdich-2018-sanskrit,sandhan-etal-2022-translist}. Ancient Chinese studies similarly explore BiLSTM--CRF analysis and historical pretraining \citep{cheng-etal-2020-integration,chang-etal-2022-automatic}, while EvaHan identifies blind-text and OOV degradation \citep{li-etal-2022-first}. Tangut intensifies these challenges, motivating data-efficient integration of expert, lexical, and unlabeled resources.

\textbf{Tangut information processing.} Tangut computation has focused mainly on digital infrastructure: fonts, Web processing, and databases \citep{nakajima-etal-1996-tangut,liu-du-2008-tangut-processing,liu-2010-tangut-font-database} followed by UCS standardization \citep{west-etal-2014-tangut}. Later work developed document recognition and character databases \citep{liu-2012-tangut-document-recognition,zhang-han-2017-tangut-recognition,ma-etal-2022-tangut-recognition}, manuscript text analysis \citep{shu-etal-2025-tangut-text-analysis}, and Tangut--Chinese translation \citep{zheng-yu-2025-tangut-translation,zheng-etal-2025-tangut-ocr-mt}. Word segmentation and contextual representation remain largely unexplored \citep{sun-xixia}.

\section{Task Definition and Data}

\subsection{Task Formulation}

Given a line of \(n\) Tangut characters
$\textbf{x}=(c_1,c_2,\ldots,c_n)$, 
the goal of word segmentation is to identify its word boundaries. We formulate the task as character-level BIES sequence labeling, where \(B\), \(I\), and \(E\) denote the beginning, inside, and end of a multi-character word, and \(S\) denotes a single-character word. The model assigns each character a label \(y_i\in\{B,I,E,S\}\) and predicts the highest-scoring valid sequence:
\[
\hat{\textbf{y}}
=\arg\max_{\textbf{y}\in\mathcal{Y}(\textbf{x})}
p_\theta(\textbf{y}\mid\textbf{x}),
\]
where \(\mathcal{Y}(\textbf{x})\) is the set of valid BIES sequences. The predicted labels are then converted into non-overlapping word spans. Evaluation is span-based: a predicted word is considered correct only when its boundaries exactly match the expert annotation.

\subsection{Data Resources and Distribution}

Multiple Tangut specialists independently segmented and labeled the texts using context, reconstructed readings, translations, and traditional lexicons. Their analyses were compared and jointly reviewed to produce the reference annotation. The corpus contains two genres: Buddhist scripture and secular writing. The religious portion is drawn from fascicle 68 of the \textit{Mahāratnakūṭa-sūtra}, whereas the secular portion comes from the encyclopaedic \textit{Leilin}, which covers diverse topics and linguistic styles. The genres differ substantially in size and lexical distribution (Table~\ref{tab:corpus-statistics}).\footnote{More details about corpus appear in Appendix~\ref{sec:appendixA1}.}. The total number of word types is calculated after deduplication across the two categories. Secular documents account for approximately 91.5\% of all segments, resulting in a substantial domain imbalance. 

\begin{table}[htbp]
\centering
\small
\begin{tabular}{lrrr}
\hline
\textbf{Category} & \textbf{Segments} & \textbf{Tokens} & \textbf{Types} \\
\hline
Buddhist scriptures & 234   & 3717  & 769   \\
Secular documents   & 2516 & 28176 & 4046 \\
\hline
Total               & 2750 & 31893 & 4433 \\
\hline
\end{tabular}
\caption{Statistics of the expert-annotated Tangut corpus.}
\label{tab:corpus-statistics}
\end{table}

Our structured lexicon contains word forms, pronunciations, definitions, and entry-level metadata. We discard single-character entries and collapse duplicate forms, yielding 19,290 multi-character candidates, of which 17,082 are two-character words (88.6\%). The lexicon has high coverage of the annotated corpus: 83.1\% of corpus word types and 95.9\% of word tokens are recorded in it. From the lexicon side, however, only 14.2\% of entries exactly match an annotated word, while only 23.3\% occur as substrings in the corpus. This asymmetry indicates that the lexicon covers most corpus tokens but contains many entries that cannot be verified using the available annotation. Therefore, it should be treated as \textbf{uncertain lexical evidence} rather than a direct source of deterministic boundaries.

The unlabeled data are extracted from the Tangut lines of four-line aligned materials. Organized by manuscript image, the collection contains 663 records, 46 title-level units, and approximately 318,000 Tangut characters. Although both collections contain the \textit{Mahāratnakūṭa-sūtra}, the annotated material is from fascicle 68, which is absent from the unlabeled collection. A normalized overlap audit finds no duplicated annotated segment of three or more characters and only limited short-$n$-gram overlap (Appendix~\ref{sec:overlap-audit}). The Tangut sequences are only used for distributional statistics and pretraining.

\section{Layered Framework and Setup}

We organize Tangut word segmentation around supervised, lexical, and distributional knowledge. The supervised layer learns word boundaries; the lexical layer encodes overlapping dictionary candidates; and the distributional layer exploits unlabeled text through corpus statistics, static character embeddings, or contextual pretraining. Except for dictionary-matching baselines, all models use character-level BIES tagging with CRF decoding.

\subsection{Unified Formulation}

Given a character sequence $\textbf{x}=(c_1,\ldots,c_n)$, a model first computes an emission score vector $\textbf{s}_i$ over the four BIES labels at each position $i$. The score of a label sequence $\textbf{y}$ is
$$
\text{Score}(\textbf{x},\textbf{y})
=
\sum_{i=1}^{n}s_i(y_i)
+
\sum_{i=2}^{n}A_{y_{i-1},y_i},
$$
where $A$ is the transition matrix learned by the CRF. Decoding selects the highest-scoring valid sequence:
$$
\hat{\textbf{y}}
=
\arg\max_{\textbf{y}\in\mathcal{Y}(\textbf{x})}
\text{Score}(\textbf{x},\textbf{y}),
$$
where $\mathcal{Y}(\textbf{x})$ denotes the set of valid BIES sequences.

The models differ primarily in how they compute the emission scores. In the linear CRF, these scores are linear combinations of discrete character features and real-valued external features. Neural models first derive a character representation $\textbf{h}_i$ and then combine it with lexical features $\textbf{z}^{\mathrm{lex}}_i$ and distributional features $\textbf{z}^{\mathrm{dist}}_i$:
$$
\textbf{s}_i
=
g_\theta
\left(
\textbf{h}_i,
\textbf{z}^{\mathrm{lex}}_i,
\textbf{z}^{\mathrm{dist}}_i
\right).
$$

This shared formulation isolates two sources of variation across models: how characters are represented and how external knowledge is incorporated.

\subsection{Supervised Segmentation Layer}

\textbf{Dictionary matching.}
As a non-parametric baseline, we apply maximum matching to determine word boundaries directly. We construct matching lexicons from the training vocabulary, the external dictionary, and their union, and compare them by bidirectional maximum matching (BMM). These baselines provide a reference for how far lexical coverage alone can support segmentation.

The \textbf{linear CRF} uses the current character, a two-character context window on each side, adjacent character bigrams, character-type indicators for digits, letters, and punctuation, and whether the current character occurs as a single-character word in the training data. It serves both as our primary supervised baseline and as the basic model for evaluating the data efficiency of explicit external knowledge.

The \textbf{BiLSTM--CRF} maps each character to a trainable embedding and encodes its left and right contexts with a bidirectional LSTM. A linear layer then produces BIES emission scores, followed by CRF decoding. Lexicon-lattice and corpus-statistical features can be concatenated with the character embeddings at the encoder input.

To separate the effect of the Transformer architecture from that of unlabeled-text pretraining, we include a randomly initialized \textbf{Transformer--CRF}. It uses the same encoder architecture as TangutEncoder but receives no masked-language-model pretraining; all parameters are learned solely from the annotated segmentation data.

\subsection{Lexical Knowledge Layer}

Multiple dictionary entries may overlap at the same character position.
Maximum matching resolves this ambiguity prematurely by retaining only
one candidate. We instead preserve all matched spans in an \textbf{aggregated
lexicon-lattice} representation. For each character, we derive 11 binary
features according to its position in a candidate and the candidate
length:$B2,B3,B4,B5{+}; I3,I4,I5{+}$ and $E2,E3,E4,E5{+}.$
Here, $B$, $I$, and $E$ indicate that the character occurs at the
beginning, inside, or at the end of a candidate. Entries of length five
or greater are grouped into the $5{+}$ category. The $I2$ feature is
omitted because a two-character word has no internal position.
Single-character entries are handled by the basic character features.

Dictionary entries do not necessarily conform to the segmentation
standard of the annotated corpus. We therefore estimate \textbf{entry-specific reliability}. For an entry $w$, $\operatorname{occ}(w)$ denotes the
number of times it occurs as a substring in the training corpus, while
$\operatorname{hit}(w)$ counts occurrences whose spans exactly match
gold words. Reliability is estimated using a length-group prior:
\[
r(w)
=
\frac{
\operatorname{hit}(w)+\kappa p_{g(w)}
}{
\operatorname{occ}(w)+\kappa
},
\]
where $\kappa=5$ controls the smoothing strength and $p_{g(w)}$ is the
prior reliability of the length group containing $w$:
\[
p_g
=
\frac{
\sum_{w\in g}\operatorname{hit}(w)+1
}{
\sum_{w\in g}\operatorname{occ}(w)+2
}.
\]
Observed entries use $r(w)$, whereas entries unseen in the training
corpus fall back to $p_g$. At each character position, we take the
maximum reliability among candidates in which the character occurs at
a beginning, internal, or ending position. This produces three
reliability features for observed entries and three prior features for
unseen entries.

We additionally derive three binary \textbf{metadata features} from the
dictionary: whether any candidate covering the
current character has a semantic-gloss entry (\texttt{has\_yi}), a
phonetic entry (\texttt{has\_yin}), or a book-title label
(\texttt{has\_book\_title}). Each feature is aggregated by logical OR:
it is set to 1 if at least one covering candidate has the
corresponding attribute. These features distinguish different sources
of lexicographic evidence: including semantically glossed entries,
phonetic or transliterated forms, and multi-character titles
, without treating any of them as a deterministic boundary.

Therefore, the complete representation of the dictionary contains 20 dimensions: 11 lattice indicators, 3 observed-entry reliability features,
3 unseen-entry prior features, and 3 lexicographic metadata
features, as summarized in Table~\ref{tab:lexicon-features}.

\begin{table}[h]
\centering
\small
\begin{tabular}{llr}
\hline
\textbf{Symbol} & \textbf{Feature group} & \textbf{Dim.} \\
\hline
\textbf{BIE} & Lexicon-lattice indicators & 11 \\
\textbf{R}   & Reliability of observed entries & 3 \\
\textbf{P}   & Prior for unseen entries & 3 \\
\textbf{M}   & Lexicographic metadata & 3 \\
\hline
$\mathrm{Dict}_{\mathrm{all}}$ & Full lexicon representation & 20 \\
\hline
\end{tabular}
\caption{Composition of the 20-dimensional lexicon representation.}
\label{tab:lexicon-features}
\end{table}

Because reliability estimation uses gold boundaries, computing it
directly on the training instances would cause information leakage.
We therefore generate training features out of fold. Within the
training portion of each outer fold, an additional five-fold split is
performed, and reliability for each inner fold is estimated using only
the other four folds. Validation and test features use statistics
estimated from the complete outer training portion. Consequently, no
line contributes its own gold boundaries to its reliability
features. The three metadata features require no such treatment because
they are intrinsic dictionary attributes and do not depend on the
annotated corpus.

\subsection{Distributional Knowledge Layer}

We exploit unlabeled Tangut text from the four-line parallel materials in three ways: explicit distributional features, static character embeddings, and contextual pretraining.

\subsubsection{Explicit Distributional Features}

For each gap between adjacent characters $a$ and $b$, we compute four types of statistics: \textbf{log bigram frequency, character association strength, the right-neighbor entropy of $a$, and the left-neighbor entropy of $b$}. Bigram frequency is defined as
$$
\text{Freq}(a,b)
=
\log\left(1+\text{count}(a,b)\right).
$$

We use discounted pointwise mutual information (dPMI) as the association measure:
$$
\text{dPMI}(a,b)
=
\max\left(
0,\,
\log
\frac{
f(a,b)-0.5
}{
f(a)\cdot f(b)
}
+
\log N
\right),
$$
where $N$ is the total number of adjacent character pairs. Alternative association measures are discussed in the Appendix~\ref{sec:appendixC}.

Neighbor entropy measures the diversity of a character's local contexts. The right-neighbor entropy of $a$ is
$$
H_{\mathrm{right}}(a)
=
-\sum_b p(b\mid a)\log p(b\mid a),
$$
with left-neighbor entropy defined analogously. For a character at position $i$, features are extracted from both its left gap $(c_{i-1},c_i)$ and right gap $(c_i,c_{i+1})$. Each gap contributes one frequency feature, one association feature, and two entropy features, yielding an eight-dimensional representation in total. (Table~\ref{tab:distributional-features})

\begin{table}[h]
\centering
\small
\begin{tabular}{llr}
\hline
\textbf{Symbol} & \textbf{Feature group} & \textbf{Dim.} \\
\hline
\textbf{Freq} & Bigram frequency & 2 \\
\textbf{Coo} & Character association & 2 \\
\textbf{Ent} & Neighbor entropy & 4 \\
\hline
$\mathrm{Dist}_{\mathrm{all}}$ & Full distributional representation & 8 \\
\hline
\end{tabular}
\caption{Explicit distributional features extracted from unlabeled text.}
\label{tab:distributional-features}
\end{table}

Unattested bigrams receive zero frequency and association values, and
characters absent from the unlabeled corpus receive zero entropy.

\subsubsection{Static Char2Vec Representations}

We train a Skip-gram model on the unlabeled Tangut text, treating each character as a basic unit. We refer to this model as Char2Vec to distinguish this character-level training from word-level Word2Vec. The learned vectors initialize the character embeddings of the Transformer and remain trainable during segmentation. Char2Vec provides each character with a distributional initialization, but this representation is context independent.

\subsubsection{Contextual Pretraining with TangutEncoder}

To learn context-dependent character representations, we develop
\textbf{TangutEncoder (TEnc)}, a compact character-level BERT-style
encoder tailored to the Tangut writing system and the limited scale of
the available unlabeled corpus. Each Tangut character is treated as a
single token, avoiding any prerequisite word segmentation. The model
uses a Tangut-specific character vocabulary and consists of character
and positional embeddings followed by three Transformer encoder layers,
with approximately 2.5 million parameters. Full architectural details are provided in
Appendix~\ref{sec:appendixB}.

TangutEncoder is pretrained with character-level masked language modeling that combines single-character and contiguous-span
masking. Approximately 15\% of character positions are selected using a mixture of single-character masking and contiguous spans of two to four characters. 80\% of selected positions are replaced with the mask token, 10\% with random characters, and 10\% remain unchanged. 

\subsection{Knowledge Fusion}

External knowledge is incorporated at different stages according to the structure of each model. In the \textbf{linear CRF}, lexicon-lattice and corpus-statistical features are added directly as real-valued feature functions alongside the basic character features. In the \textbf{BiLSTM--CRF}, external features are concatenated with character embeddings at the encoder input and jointly contextualized by the BiLSTM. During training, lexicon features are randomly dropped. In \textbf{TangutEncoder}, external features are incorporated after contextual encoding. Lexicon and corpus-statistical features can be transformed into a task-specific representation through a linear layer, ReLU activation, and dropout, and then concatenated with the Transformer output. The combined representation is passed through another dropout layer and a linear projection to produce BIES emission scores. 

\subsection{Experimental Protocol}

\textbf{Data splits.}
We conduct line-level five-fold cross-validation stratified by genre. In each round, one fold is held out for testing, while one ninth of the remaining data is used for validation. Because both works occur across folds, this protocol evaluates within-source generalization to held-out, thematically diverse passages rather than transfer to unseen documents. Reliability features are generated by an additional five-fold out-of-fold procedure within each training partition.

\textbf{Evaluation metrics.}
Segmentation is evaluated by exact word-span matching. Let $P$ be the set of predicted spans and $G$ the set of gold spans. A word is counted as correct only when both of its boundaries exactly match a gold word. Precision, recall, and F$_1$ are defined as
$$
\text{Precision}
=
\frac{|P\cap G|}{|P|},
\ 
\text{Recall}
=
\frac{|P\cap G|}{|G|},
$$
$$
F_1
=
\frac{
2\cdot\text{Precision}\cdot\text{Recall}
}{
\text{Precision}+\text{Recall}
}.
$$

We define a gold test word as out of vocabulary (OOV) when its surface form is absent from the labeled training partition of the current fold; all other words are in vocabulary (IV). An OOV word may still occur in the external lexicon or unlabeled corpus. Thus, OOV recall measures generalization beyond the labeled training vocabulary rather than to forms unseen by every system component:
$$
\text{OOV\text{-}R}
=
\frac{
\text{correctly segmented OOV words}
}{
\text{gold OOV words}
},
$$
$$
\text{IV\text{-}R}
=
\frac{
\text{correctly segmented IV words}
}{
\text{gold IV words}
}.
$$

Across the five held-out folds, 53.0\% of labeled-training OOV tokens are covered by the external lexicon and 34.8\% occur in the unlabeled corpus; these sets overlap, while 24.8\% occur in neither resource (Appendix~\ref{sec:overlap-audit}).

In addition to aggregate results, we report performance separately on religious and secular texts. All metrics are summarized as the mean and standard deviation across the five outer folds.

\textbf{Training and model selection.}
The linear CRF is optimized with L-BFGS and L1/L2 regularization. For the BiLSTM--CRF and Transformer--CRF models, the learning rate is reduced when validation loss stops improving, and the checkpoint with the lowest validation loss is restored after early stopping. Downstream training of TangutEncoder proceeds in two stages. We first freeze the encoder and train only the feature projections, segmentation head and CRF. Then we unfreeze the encoder and jointly fine-tune all parameters.

All experiments use fixed random seeds and deterministic CuDNN settings within each training environment. Models are compared under identical data splits, lexical resources, and evaluation procedures. Full settings are provided in Appendix~\ref{sec:appendixB}.


\begin{table*}[h]
\centering
\footnotesize
\setlength{\tabcolsep}{3.8pt}
\begin{tabular}{lccccc}
\toprule
Model & P & R & F$_1$ & OOV-R & IV-R \\
\midrule
\multicolumn{6}{l}{\textit{Supervised baselines (RQ1)}} \\
Dict-corpus
& $0.845{\pm}0.005$ & $0.890{\pm}0.003$ & $0.867{\pm}0.004$
& $0.219{\pm}0.014$ & $\textbf{0.954}{\pm}0.002$ \\
CRF
& $0.880{\pm}0.004$ & $0.888{\pm}0.003$ & $0.884{\pm}0.003$
& $0.415{\pm}0.026$ & $0.933{\pm}0.003$ \\
BiLSTM--CRF
& $0.862{\pm}0.008$ & $0.873{\pm}0.008$ & $0.868{\pm}0.008$
& $0.472{\pm}0.014$ & $0.912{\pm}0.006$ \\
Transformer-Random
& $0.848{\pm}0.008$ & $0.863{\pm}0.010$ & $0.855{\pm}0.009$
& $0.519{\pm}0.020$ & $0.896{\pm}0.010$ \\
\midrule
\multicolumn{6}{l}{\textit{Lexicon-lattice ablation (RQ2)}} \\
CRF+BIE
& $0.898{\pm}0.004$ & $0.897{\pm}0.002$ & $0.897{\pm}0.003$
& $0.489{\pm}0.015$ & $0.936{\pm}0.002$ \\
\quad + BIE + R
& $0.902{\pm}0.005$ & $0.898{\pm}0.003$ & $0.900{\pm}0.004$
& $0.473{\pm}0.021$ & $0.939{\pm}0.002$ \\
\quad + BIE + R + P
& $0.900{\pm}0.003$ & $0.900{\pm}0.005$ & $0.900{\pm}0.004$ 
& $0.463{\pm}0.020$ & $0.941{\pm}0.003$ \\
\quad + $\text{Dict}_{\text{all}}$
& $0.903{\pm}0.005$ & $0.902{\pm}0.003$ & $0.902{\pm}0.004$
& $0.467{\pm}0.021$ & $0.944{\pm}0.002$ \\
\midrule
\multicolumn{6}{l}{\textit{Explicit distributional features (RQ3)}} \\
CRF+ $\text{Dict}_{\text{all}}$ + Freq
& $0.905{\pm}0.003$ & $0.904{\pm}0.003$ & $0.904{\pm}0.003$
& $0.483{\pm}0.011$ & $0.944{\pm}0.002$ \\
\quad + $\text{Dict}_{\text{all}}$ + Freq + Coo
& $0.906{\pm}0.003$ & $0.905{\pm}0.003$ & $0.905{\pm}0.003$
& $0.486{\pm}0.009$ & $0.945{\pm}0.001$ \\
\quad + $\text{Dict}_{\text{all}}$ + $\text{Dist}_{\text{all}}$
& $\textbf{0.907}{\pm}0.004$
& $0.906{\pm}0.004$
& $0.907{\pm}0.004$
& $0.491{\pm}0.016$
& $0.947{\pm}0.002$ \\
\midrule
\multicolumn{6}{l}{\textit{Representation learning and fusion (RQ4--RQ5)}} \\
Transformer-Char2Vec
& $0.844{\pm}0.005$ & $0.860{\pm}0.012$ & $0.852{\pm}0.008$
& $0.491{\pm}0.010$ & $0.895{\pm}0.012$ \\
TangutEncoder
& $0.881{\pm}0.004$ & $0.885{\pm}0.002$ & $0.883{\pm}0.003$
& $0.570{\pm}0.011$ & $0.915{\pm}0.004$ \\
TangutEncoder + $\text{Dict}_{\text{all}}$
& $0.902{\pm}0.003$ & $0.913{\pm}0.006$ & $0.907{\pm}0.004$
& $0.606{\pm}0.014$ & $0.942{\pm}0.005$ \\
TangutEncoder + $\text{Dict}_{\text{all}}$ + $\text{Dist}_{\text{all}}$
& $\textbf{0.905}{\pm}0.003$ & $\textbf{0.916}{\pm}0.004$ & $\textbf{0.911}{\pm}0.003$
& $\textbf{0.608}{\pm}0.014$ & $0.946{\pm}0.003$ \\
\bottomrule
\end{tabular}
\caption{Overall segmentation results under five-fold cross-validation.}
\label{tab:overall-results}
\end{table*}

\begin{table*}[h]
\centering
\footnotesize
\setlength{\tabcolsep}{4.2pt}
\begin{tabular}{lcccccc}
\toprule
& \multicolumn{3}{c}{Secular} & \multicolumn{3}{c}{Religious} \\
\cmidrule(lr){2-4}\cmidrule(lr){5-7}
Model & F$_1$ & OOV-R & IV-R & F$_1$ & OOV-R & IV-R \\
\midrule
Dict-corpus
& $0.869{\pm}0.004$ & $0.229{\pm}0.017$ & $\textbf{0.956}{\pm}0.003$
& $0.855{\pm}0.019$ & $0.132{\pm}0.028$ & $0.936{\pm}0.010$ \\
CRF
& $0.887{\pm}0.010$ & $0.430{\pm}0.035$ & $0.934{\pm}0.007$
& $0.839{\pm}0.011$ & $0.262{\pm}0.067$ & $0.909{\pm}0.009$ \\
BiLSTM--CRF
& $0.875{\pm}0.007$ & $0.498{\pm}0.016$ & $0.917{\pm}0.006$
& $0.810{\pm}0.021$ & $0.244{\pm}0.028$ & $0.875{\pm}0.021$ \\
Transformer-Random
& $0.868{\pm}0.008$ & $0.545{\pm}0.018$ & $0.906{\pm}0.010$
& $0.760{\pm}0.014$ & $0.280{\pm}0.060$ & $0.816{\pm}0.017$ \\
CRF+$\text{Dict}_{\text{all}}$
& $0.904{\pm}0.005$ & $0.481{\pm}0.024$ & $0.942{\pm}0.004$
& $0.872{\pm}0.004$ & $0.301{\pm}0.058$ & $0.938{\pm}0.008$ \\
CRF+$\text{Dict}_{\text{all}}$ + $\text{Dist}_{\text{all}}$
& $0.910{\pm}0.005$ & $0.506{\pm}0.022$ & $0.947{\pm}0.002$
& $\textbf{0.881}{\pm}0.005$ & $0.355{\pm}0.060$ & $\textbf{0.939}{\pm}0.008$ \\
Transformer-Char2Vec
& $0.866{\pm}0.008$ & $0.514{\pm}0.011$ & $0.908{\pm}0.012$
& $0.749{\pm}0.016$ & $0.284{\pm}0.033$ & $0.800{\pm}0.022$ \\
TEnc
& $0.892{\pm}0.002$ & $0.587{\pm}0.016$ & $0.922{\pm}0.004$
& $0.817{\pm}0.016$ & $0.422{\pm}0.039$ & $0.863{\pm}0.015$ \\
TEnc + $\text{Dict}_{\text{all}}$
& $0.915{\pm}0.004$ & $0.633{\pm}0.023$ & $0.946{\pm}0.006$
& $0.849{\pm}0.018$ & $0.362{\pm}0.065$ & $0.908{\pm}0.007$ \\
TEnc + $\text{Dict}_{\text{all}}$ + $\text{Dist}_{\text{all}}$
& $\textbf{0.917}{\pm}0.003$ & $\textbf{0.634}{\pm}0.022$ & $0.949{\pm}0.004$
& $0.863{\pm}0.015$ & $\textbf{0.385}{\pm}0.058$ & $0.920{\pm}0.010$ \\
\bottomrule
\end{tabular}
\caption{Mean performance by document genre.}
\label{tab:genre-results}
\end{table*}

\section{Results}

Table~\ref{tab:overall-results} summarizes the five research questions. The OOV rate is approximately 0.087 across folds and is therefore omitted. Table~\ref{tab:genre-results} reports genre-specific results for representative systems; full genre-level ablations are provided in Appendix~\ref{sec:appendixC}. Standard deviations reflect variation across data folds rather than repeated random initializations.

\subsection{Supervised Baselines under Limited Annotation}

\paragraph{RQ1: How do basic segmenters perform with limited annotation?}
The linear CRF is the strongest supervised baseline overall.
Corpus-dictionary matching recognizes IV words well but generalizes
poorly beyond the labeled vocabulary. Neural models improve OOV recall;
the random Transformer gives the highest OOV recall but the lowest
overall F$_1$. Thus, explicit local features remain more data-efficient
for overall segmentation.

All learned baselines perform worse on religious than on secular texts,
consistent with the substantial imbalance between the two genres.

\subsection{Reliability-Calibrated Lexicon Features}

\paragraph{RQ2: Do lexicon-lattice features and reliability calibration help?}
BIE lattice features provide the largest lexical gain, confirming that
overlapping dictionary candidates supply useful boundary evidence.
Reliability estimates for observed entries further improve precision
and IV recognition, although OOV recall decreases slightly. Priors for
unseen entries maintain the overall improvement. Completing the representation with dictionary
metadata yields the strongest lexicon-only CRF, suggesting that
distinguishing semantically glossed, phonetic, and book-title entries
helps the model assess heterogeneous candidates. Overall, the lattice
provides coverage, while reliability and metadata help control
mismatches between dictionary entries and corpus boundaries.

\subsection{Explicit Distributional Features}

\paragraph{RQ3: How does unlabeled text help the linear CRF?}
Statistics extracted from unlabeled text provide additional gains beyond
the complete 20-dimensional lexicon representation. Bigram frequency
contributes the largest initial improvement, while character association
and neighbor entropy add complementary evidence about local cohesion and
potential boundaries. The gains extend to both IV and OOV recognition
and are particularly visible for religious texts. Thus, dictionary
evidence and corpus distribution capture complementary aspects of
lexical structure.

\subsection{Static versus Contextual Representations}

\paragraph{RQ4: How do static and contextual representations differ?}
Char2Vec does not improve consistently over random initialization; its
slightly lower mean lies within cross-fold variation. In contrast, MLM
pretraining substantially improves F$_1$, OOV recall, and religious-text
performance by training the entire contextual encoder rather than only
providing static character embeddings.

\subsection{Comparison of Knowledge-Fusion Strategies}

\paragraph{RQ5: Are lexical and distributional knowledge complementary?}
Adding the complete lexicon representation to TangutEncoder markedly
improves both IV and OOV recognition, showing that explicit lexical
knowledge remains valuable after contextual pretraining. Injecting
corpus statistics provides a smaller additional numerical gain,
indicating that MLM captures much, but not all, of the available local
distributional information.

The complete systems remain complementary. TangutEncoder has the highest
overall and secular-text F$_1$ and OOV recall, whereas the CRF retains
slightly higher overall IV recall and performs better on religious
texts. Contextual pretraining therefore chiefly improves generalization
beyond the labeled vocabulary; explicit features remain stable in the
smallest genre.

\section{Preliminary Extension to POS Tagging}

We conduct a preliminary experiment on joint word segmentation and part-of-speech tagging by replacing the four BIES labels with joint boundary--POS labels, such as \texttt{B-NOUN} and \texttt{E-NOUN}. We apply this formulation to both the linear CRF and BiLSTM--CRF, with and without the full lexical features introduced above. The predicted joint sequence is first converted into word spans and then into POS labels. POS accuracy is calculated only over words whose predicted spans exactly match the gold segmentation.

The joint CRF reaches approximately 0.88 conditional POS accuracy on correctly segmented words. Lexical features improve segmentation, but their contribution to POS is less stable, suggesting that dictionary evidence is more directly informative about boundaries than grammatical categories. Because the POS inventory remains under revision (Appendix~\ref{pos}), these results are preliminary.

\section{Cross-Document Evaluation}
\label{sec:unseen-evaluation}

We further evaluate cross-document generalization on the independent test set (Appendix~\ref{sec:independent-test}). Each model is trained on all 2,750 original segments using fixed configurations, without tuning on the 479 test segments from five additional works.

\begin{table}[t]
\centering
\small
\setlength{\tabcolsep}{5pt}
\begin{tabular}{lc}
\toprule
\textbf{Model} & \textbf{F$_1$} \\
\midrule
CRF                                      & 0.686 \\
CRF + $\mathrm{Dict}_{\mathrm{all}}$     & 0.726 \\
CRF + $\mathrm{Dict}_{\mathrm{all}}$
    + $\mathrm{Dist}_{\mathrm{all}}$     & 0.747 \\
\midrule
Transformer-Random                              & 0.642 \\
Transformer-Char2Vec                            & 0.652 \\
TangutEncoder                                & 0.722 \\
TangutEncoder + $\mathrm{Dict}_{\mathrm{all}}$
    + $\mathrm{Dist}_{\mathrm{all}}$     & 0.754 \\
\bottomrule
\end{tabular}
\caption{Segmentation results on five unseen works. Full results are provided in Appendix~\ref{sec:appendixC4}.}
\label{tab:unseen-results}
\end{table}

All models perform substantially worse than under within-source
cross-validation, indicating considerable lexical and stylistic variation
across documents. Nevertheless, the main findings remain consistent.
Adding the lexicon lattice and explicit distributional features
progressively improves the linear CRF, while MLM pretraining provides a
much larger transfer gain than random or Char2Vec initialization.
Combining TangutEncoder with lexical and distributional knowledge yields
the strongest standalone segmenter. Overall, external knowledge and contextual
pretraining remain complementary under cross-document evaluation.

\section{Conclusion and Future Work}

This study presents the first systematic investigation of automatic
Tangut word segmentation, combining expert annotation, traditional
lexicons, and unlabeled text in a BIES--CRF framework. Lexicon-lattice
features and corpus statistics provide efficient boundary evidence,
while contextual MLM pretraining outperforms random and Char2Vec
initialization. The best main system achieves 0.911 mean F$_1$ in
within-source evaluation, and the best standalone segmenter reaches
0.754 on five additional works, showing useful but limited
cross-document generalization.

Future work will expand annotation across documents, genres, and
periods, refine the POS inventory and guidelines, and improve expert
consistency assessment. We will seek permission to release more
annotation or provide controlled access alongside code and models.
Larger digitized and parallel corpora will support further pretraining,
weak supervision, significance testing, and downstream evaluation.


\section*{Limitations}

The development corpus remains small and genre-imbalanced. Although the independent test set provides document-level unseen, within-domain evaluation, it contains only five additional works and does not establish generalization across the full range of Tangut genres and historical periods. Moreover, limited overlap with the unlabeled corpus means that this evaluation is not a strict zero-exposure setting. Moreover, small model differences lack multi-seed or significance testing. Finally, data-use restrictions prevent full corpus release, and the POS inventory remains under revision.

\section*{Ethical Considerations}

This study uses expert-annotated Tangut texts and digitized historical
materials for research purposes. The full annotated corpus cannot be
publicly released because of institutional and data-use restrictions.
Where permitted, the release will include code, deterministic split
generation and fold identifiers, normalization and feature scripts,
model configurations, a lexicon-processing manifest and checksum,
annotation guidelines and examples, source-location metadata, trained
models, and derived statistics. The data contain historical texts rather than personal
or sensitive information. Nevertheless, segmentation errors may
propagate to retrieval, lexical analysis, and translation, particularly
for underrepresented document genres. The resulting models should
therefore be regarded as research aids rather than substitutes for
expert philological interpretation. More broadly, this work is intended
to support the preservation, retrieval, and linguistic analysis of
Tangut cultural heritage.

\section*{Acknowledgments}

This work was supported by the Discipline Development ``Peak-Climbing Strategy'' Funding Program of the Chinese Academy of Social Sciences (Project No.~DF2023TS05) and the Key Laboratory of Linguistics, Chinese Academy of Social Sciences (Project No.~2024SYZH001).

\bibliography{custom}

@inproceedings{xue-2003-chinese,
    title = "{C}hinese Word Segmentation as Character Tagging",
    author = "Xue, Nianwen",
    booktitle = "International Journal of Computational Linguistics {\&} {C}hinese Language Processing, Volume 8, Number 1, {F}ebruary 2003: Special Issue on Word Formation and {C}hinese Language Processing",
    month = feb,
    year = "2003",
    url = "https://aclanthology.org/O03-4002/",
    pages = "29--48"
}

@inproceedings{peng-etal-2004-chinese,
    title = "{C}hinese Segmentation and New Word Detection using Conditional Random Fields",
    author = "Peng, Fuchun  and
      Feng, Fangfang  and
      McCallum, Andrew",
    booktitle = "{COLING} 2004: Proceedings of the 20th International Conference on Computational Linguistics",
    month = "aug 23–aug 27",
    year = "2004",
    address = "Geneva, Switzerland",
    publisher = "COLING",
    url = "https://aclanthology.org/C04-1081/",
    pages = "562--568"
}

@inproceedings{cai-zhao-2016-neural,
    title = "Neural Word Segmentation Learning for {C}hinese",
    author = "Cai, Deng  and
      Zhao, Hai",
    editor = "Erk, Katrin  and
      Smith, Noah A.",
    booktitle = "Proceedings of the 54th Annual Meeting of the Association for Computational Linguistics (Volume 1: Long Papers)",
    month = aug,
    year = "2016",
    address = "Berlin, Germany",
    publisher = "Association for Computational Linguistics",
    url = "https://aclanthology.org/P16-1039/",
    doi = "10.18653/v1/P16-1039",
    pages = "409--420"
}

@inproceedings{ma-etal-2018-state,
    title = "State-of-the-art {C}hinese Word Segmentation with {B}i-{LSTM}s",
    author = "Ma, Ji  and
      Ganchev, Kuzman  and
      Weiss, David",
    editor = "Riloff, Ellen  and
      Chiang, David  and
      Hockenmaier, Julia  and
      Tsujii, Jun{'}ichi",
    booktitle = "Proceedings of the 2018 Conference on Empirical Methods in Natural Language Processing",
    month = oct # "-" # nov,
    year = "2018",
    address = "Brussels, Belgium",
    publisher = "Association for Computational Linguistics",
    url = "https://aclanthology.org/D18-1529/",
    doi = "10.18653/v1/D18-1529",
    pages = "4902--4908"
}

@article{shao-etal-2018-universal,
    title = "Universal Word Segmentation: Implementation and Interpretation",
    author = "Shao, Yan  and
      Hardmeier, Christian  and
      Nivre, Joakim",
    editor = "Lee, Lillian  and
      Johnson, Mark  and
      Toutanova, Kristina  and
      Roark, Brian",
    journal = "Transactions of the Association for Computational Linguistics",
    volume = "6",
    year = "2018",
    address = "Cambridge, MA",
    publisher = "MIT Press",
    url = "https://aclanthology.org/Q18-1030/",
    doi = "10.1162/tacl_a_00033",
    pages = "421--435"
}

@inproceedings{liu-etal-2011-tibetan,
    title = "{T}ibetan Word Segmentation as Syllable Tagging Using Conditional Random Field",
    author = "Liu, Huidan  and
      Nuo, Minghua  and
      Ma, Longlong  and
      Wu, Jian  and
      He, Yeping",
    editor = "Gao, Helena Hong  and
      Dong, Minghui",
    booktitle = "Proceedings of the 25th Pacific Asia Conference on Language, Information and Computation",
    month = dec,
    year = "2011",
    address = "Singapore",
    publisher = "Institute of Digital Enhancement of Cognitive Processing, Waseda University",
    url = "https://aclanthology.org/Y11-1018/",
    pages = "168--177"
}

@inproceedings{brown-2024-improved,
    title = "Improved Neural Word Segmentation for Standard {T}ibetan",
    author = "Brown, Collin J.",
    editor = "Ojha, Atul Kr.  and
      Ahmadi, Sina  and
      Cinkov{\'a}, Silvie  and
      Fransen, Theodorus  and
      Liu, Chao-Hong  and
      McCrae, John P.",
    booktitle = "Proceedings of the 2nd Workshop on Resources and Technologies for Indigenous, Endangered and Lesser-resourced Languages in Eurasia (EURALI) @ LREC-COLING 2024",
    month = may,
    year = "2024",
    address = "Torino, Italia",
    publisher = "ELRA and ICCL",
    url = "https://aclanthology.org/2024.eurali-1.2/",
    pages = "12--17"
}

@inproceedings{wang-xu-2017-convolutional,
    title = "Convolutional Neural Network with Word Embeddings for {C}hinese Word Segmentation",
    author = "Wang, Chunqi  and
      Xu, Bo",
    editor = "Kondrak, Greg  and
      Watanabe, Taro",
    booktitle = "Proceedings of the Eighth International Joint Conference on Natural Language Processing (Volume 1: Long Papers)",
    month = nov,
    year = "2017",
    address = "Taipei, Taiwan",
    publisher = "Asian Federation of Natural Language Processing",
    url = "https://aclanthology.org/I17-1017/",
    pages = "163--172"
}

@inproceedings{htay-murthy-2008-myanmar,
    title = "{M}yanmar Word Segmentation using Syllable level Longest Matching",
    author = "Htay, Hla Hla  and
      Murthy, Kavi Narayana",
    booktitle = "Proceedings of the 6th Workshop on {A}sian Language Resources",
    year = "2008",
    url = "https://aclanthology.org/I08-7006/"
}

@inproceedings{liu-etal-2014-domain,
    title = "Domain Adaptation for {CRF}-based {C}hinese Word Segmentation using Free Annotations",
    author = "Liu, Yijia  and
      Zhang, Yue  and
      Che, Wanxiang  and
      Liu, Ting  and
      Wu, Fan",
    editor = "Moschitti, Alessandro  and
      Pang, Bo  and
      Daelemans, Walter",
    booktitle = "Proceedings of the 2014 Conference on Empirical Methods in Natural Language Processing ({EMNLP})",
    month = oct,
    year = "2014",
    address = "Doha, Qatar",
    publisher = "Association for Computational Linguistics",
    url = "https://aclanthology.org/D14-1093/",
    doi = "10.3115/v1/D14-1093",
    pages = "864--874"
}

@article{
  Zhang_Liu_Fu_2018, 
  title={Neural Networks Incorporating Dictionaries for Chinese Word Segmentation}, 
  volume={32}, 
  url={https://ojs.aaai.org/index.php/AAAI/article/view/11959}, 
  DOI={10.1609/aaai.v32i1.11959}, 
  abstractNote={ &amp;lt;p&amp;gt; In recent years, deep neural networks have achieved significant success in Chinese word segmentation and many other natural language processing tasks. Most of these algorithms are end-to-end trainable systems and can effectively process and learn from large scale labeled datasets. However, these methods typically lack the capability of processing rare words and data whose domains are different from training data. Previous statistical methods have demonstrated that human knowledge can provide valuable information for handling rare cases and domain shifting problems. In this paper, we seek to address the problem of incorporating dictionaries into neural networks for the Chinese word segmentation task. Two different methods that extend the bi-directional long short-term memory neural network are proposed to perform the task. To evaluate the performance of the proposed methods, state-of-the-art supervised models based methods and domain adaptation approaches are compared with our methods on nine datasets from different domains. The experimental results demonstrate that the proposed methods can achieve better performance than other state-of-the-art neural network methods and domain adaptation approaches in most cases.&amp;lt;br /&amp;gt; &amp;lt;/p&amp;gt; }, 
  number={1}, 
  journal={Proceedings of the AAAI Conference on Artificial Intelligence}, 
  author={Zhang, Qi and Liu, Xiaoyu and Fu, Jinlan}, 
  year={2018}, 
  month={Apr.} }

@inproceedings{higashiyama-etal-2019-incorporating,
    title = "Incorporating Word Attention into Character-Based Word Segmentation",
    author = "Higashiyama, Shohei  and
      Utiyama, Masao  and
      Sumita, Eiichiro  and
      Ideuchi, Masao  and
      Oida, Yoshiaki  and
      Sakamoto, Yohei  and
      Okada, Isaac",
    editor = "Burstein, Jill  and
      Doran, Christy  and
      Solorio, Thamar",
    booktitle = "Proceedings of the 2019 Conference of the North {A}merican Chapter of the Association for Computational Linguistics: Human Language Technologies, Volume 1 (Long and Short Papers)",
    month = jun,
    year = "2019",
    address = "Minneapolis, Minnesota",
    publisher = "Association for Computational Linguistics",
    url = "https://aclanthology.org/N19-1276/",
    doi = "10.18653/v1/N19-1276",
    pages = "2699--2709"
}

@article{higashiyama-etal-2020-auxiliary,
  title={Auxiliary Lexicon Word Prediction for Cross-Domain Word Segmentation},
  author={Shohei Higashiyama and Masao Utiyama and Yuji Matsumoto and Taro Watanabe and Eiichiro Sumita},
  journal={Journal of Natural Language Processing},
  volume={27},
  number={3},
  pages={573-598},
  year={2020},
  doi={10.5715/jnlp.27.573}
}

@inproceedings{huang-etal-2021-lexicon-based,
    title = "Lexicon-Based Graph Convolutional Network for {C}hinese Word Segmentation",
    author = "Huang, Kaiyu  and
      Yu, Hao  and
      Liu, Junpeng  and
      Liu, Wei  and
      Cao, Jingxiang  and
      Huang, Degen",
    editor = "Moens, Marie-Francine  and
      Huang, Xuanjing  and
      Specia, Lucia  and
      Yih, Scott Wen-tau",
    booktitle = "Findings of the Association for Computational Linguistics: EMNLP 2021",
    month = nov,
    year = "2021",
    address = "Punta Cana, Dominican Republic",
    publisher = "Association for Computational Linguistics",
    url = "https://aclanthology.org/2021.findings-emnlp.248/",
    doi = "10.18653/v1/2021.findings-emnlp.248",
    pages = "2908--2917"
}

@inproceedings{okabe-etal-2022-weakly,
    title = "Weakly Supervised Word Segmentation for Computational Language Documentation",
    author = "Okabe, Shu  and
      Besacier, Laurent  and
      Yvon, Fran{\c{c}}ois",
    editor = "Muresan, Smaranda  and
      Nakov, Preslav  and
      Villavicencio, Aline",
    booktitle = "Proceedings of the 60th Annual Meeting of the Association for Computational Linguistics (Volume 1: Long Papers)",
    month = may,
    year = "2022",
    address = "Dublin, Ireland",
    publisher = "Association for Computational Linguistics",
    url = "https://aclanthology.org/2022.acl-long.510/",
    doi = "10.18653/v1/2022.acl-long.510",
    pages = "7385--7398"
}

@inproceedings{nigatu-etal-2024-zenos,
    title = "The {Z}eno{'}s Paradox of `Low-Resource' Languages",
    author = "Nigatu, Hellina Hailu  and
      Tonja, Atnafu Lambebo  and
      Rosman, Benjamin  and
      Solorio, Thamar  and
      Choudhury, Monojit",
    editor = "Al-Onaizan, Yaser  and
      Bansal, Mohit  and
      Chen, Yun-Nung",
    booktitle = "Proceedings of the 2024 Conference on Empirical Methods in Natural Language Processing",
    month = nov,
    year = "2024",
    address = "Miami, Florida, USA",
    publisher = "Association for Computational Linguistics",
    url = "https://aclanthology.org/2024.emnlp-main.983/",
    doi = "10.18653/v1/2024.emnlp-main.983",
    pages = "17753--17774"
}

@inproceedings{liu-best-2025-survey,
    title = "A Survey of {NLP} Progress in {S}ino-{T}ibetan Low-Resource Languages",
    author = "Liu, Shuheng  and
      Best, Michael",
    editor = "Chiruzzo, Luis  and
      Ritter, Alan  and
      Wang, Lu",
    booktitle = "Proceedings of the 2025 Conference of the Nations of the Americas Chapter of the Association for Computational Linguistics: Human Language Technologies (Volume 1: Long Papers)",
    month = apr,
    year = "2025",
    address = "Albuquerque, New Mexico",
    publisher = "Association for Computational Linguistics",
    url = "https://aclanthology.org/2025.naacl-long.396/",
    doi = "10.18653/v1/2025.naacl-long.396",
    pages = "7804--7825",
    ISBN = "979-8-89176-189-6"
}

@inproceedings{homburg-chiarcos-2016-word,
    title = "Word Segmentation for {A}kkadian Cuneiform",
    author = "Homburg, Timo  and
      Chiarcos, Christian",
    editor = "Calzolari, Nicoletta  and
      Choukri, Khalid  and
      Declerck, Thierry  and
      Goggi, Sara  and
      Grobelnik, Marko  and
      Maegaard, Bente  and
      Mariani, Joseph  and
      Mazo, Helene  and
      Moreno, Asuncion  and
      Odijk, Jan  and
      Piperidis, Stelios",
    booktitle = "Proceedings of the Tenth International Conference on Language Resources and Evaluation ({LREC}'16)",
    month = may,
    year = "2016",
    address = "Portoro{\v{z}}, Slovenia",
    publisher = "European Language Resources Association (ELRA)",
    url = "https://aclanthology.org/L16-1642/",
    pages = "4067--4074"
}

@article{luo-etal-2021-deciphering,
    title = "Deciphering Undersegmented Ancient Scripts Using Phonetic Prior",
    author = "Luo, Jiaming  and
      Hartmann, Frederik  and
      Santus, Enrico  and
      Barzilay, Regina  and
      Cao, Yuan",
    editor = "Roark, Brian  and
      Nenkova, Ani",
    journal = "Transactions of the Association for Computational Linguistics",
    volume = "9",
    year = "2021",
    address = "Cambridge, MA",
    publisher = "MIT Press",
    url = "https://aclanthology.org/2021.tacl-1.5/",
    doi = "10.1162/tacl_a_00354",
    pages = "69--81"
}

@inproceedings{krishna-etal-2017-dataset,
    title = "A Dataset for {S}anskrit Word Segmentation",
    author = "Krishna, Amrith  and
      Satuluri, Pavan Kumar  and
      Goyal, Pawan",
    editor = "Alex, Beatrice  and
      Degaetano-Ortlieb, Stefania  and
      Feldman, Anna  and
      Kazantseva, Anna  and
      Reiter, Nils  and
      Szpakowicz, Stan",
    booktitle = "Proceedings of the Joint {SIGHUM} Workshop on Computational Linguistics for Cultural Heritage, Social Sciences, Humanities and Literature",
    month = aug,
    year = "2017",
    address = "Vancouver, Canada",
    publisher = "Association for Computational Linguistics",
    url = "https://aclanthology.org/W17-2214/",
    doi = "10.18653/v1/W17-2214",
    pages = "105--114"
}

@inproceedings{hellwig-nehrdich-2018-sanskrit,
    title = "{S}anskrit Word Segmentation Using Character-level Recurrent and Convolutional Neural Networks",
    author = "Hellwig, Oliver  and
      Nehrdich, Sebastian",
    editor = "Riloff, Ellen  and
      Chiang, David  and
      Hockenmaier, Julia  and
      Tsujii, Jun{'}ichi",
    booktitle = "Proceedings of the 2018 Conference on Empirical Methods in Natural Language Processing",
    month = oct # "-" # nov,
    year = "2018",
    address = "Brussels, Belgium",
    publisher = "Association for Computational Linguistics",
    url = "https://aclanthology.org/D18-1295/",
    doi = "10.18653/v1/D18-1295",
    pages = "2754--2763"
}

@inproceedings{sandhan-etal-2022-translist,
    title = "{T}rans{LIST}: A Transformer-Based Linguistically Informed {S}anskrit Tokenizer",
    author = "Sandhan, Jivnesh  and
      Singha, Rathin  and
      Rao, Narein  and
      Samanta, Suvendu  and
      Behera, Laxmidhar  and
      Goyal, Pawan",
    editor = "Goldberg, Yoav  and
      Kozareva, Zornitsa  and
      Zhang, Yue",
    booktitle = "Findings of the Association for Computational Linguistics: EMNLP 2022",
    month = dec,
    year = "2022",
    address = "Abu Dhabi, United Arab Emirates",
    publisher = "Association for Computational Linguistics",
    url = "https://aclanthology.org/2022.findings-emnlp.513/",
    doi = "10.18653/v1/2022.findings-emnlp.513",
    pages = "6902--6912"
}

@inproceedings{cheng-etal-2020-integration,
    title = "Integration of Automatic Sentence Segmentation and Lexical Analysis of {A}ncient {C}hinese based on {B}i{LSTM}-{CRF} Model",
    author = "Cheng, Ning  and
      Li, Bin  and
      Xiao, Liming  and
      Xu, Changwei  and
      Ge, Sijia  and
      Hao, Xingyue  and
      Feng, Minxuan",
    editor = "Sprugnoli, Rachele  and
      Passarotti, Marco",
    booktitle = "Proceedings of LT4HALA 2020 - 1st Workshop on Language Technologies for Historical and Ancient Languages",
    month = may,
    year = "2020",
    address = "Marseille, France",
    publisher = "European Language Resources Association (ELRA)",
    url = "https://aclanthology.org/2020.lt4hala-1.8/",
    pages = "52--58",
    language = "eng",
    ISBN = "979-10-95546-53-5"
}

@inproceedings{li-etal-2022-first,
    title = "The First International {A}ncient {C}hinese Word Segmentation and {POS} Tagging Bakeoff: Overview of the {E}va{H}an 2022 Evaluation Campaign",
    author = "Li, Bin  and
      Yuan, Yiguo  and
      Lu, Jingya  and
      Feng, Minxuan  and
      Xu, Chao  and
      Qu, Weiguang  and
      Wang, Dongbo",
    editor = "Sprugnoli, Rachele  and
      Passarotti, Marco",
    booktitle = "Proceedings of the Second Workshop on Language Technologies for Historical and Ancient Languages",
    month = jun,
    year = "2022",
    address = "Marseille, France",
    publisher = "European Language Resources Association",
    url = "https://aclanthology.org/2022.lt4hala-1.19/",
    pages = "135--140"
}

@inproceedings{chang-etal-2022-automatic,
    title = "Automatic Word Segmentation and Part-of-Speech Tagging of {A}ncient {C}hinese Based on {BERT} Model",
    author = "Chang, Yu  and
      Zhu, Peng  and
      Wang, Chaoping  and
      Wang, Chaofan",
    editor = "Sprugnoli, Rachele  and
      Passarotti, Marco",
    booktitle = "Proceedings of the Second Workshop on Language Technologies for Historical and Ancient Languages",
    month = jun,
    year = "2022",
    address = "Marseille, France",
    publisher = "European Language Resources Association",
    url = "https://aclanthology.org/2022.lt4hala-1.20/",
    pages = "141--145"
}

@book{nakajima-etal-1996-tangut,
  author    = {Nakajima, Motoki and Imai, Kenji and Takahashi, Mariyo},
  title     = {Toward the Computational Analysis of the {Tangut} Script:
               1996 Edition},
  publisher = {Research Institute for Languages and Cultures of Asia and Africa,
               Tokyo University of Foreign Studies},
  address   = {Tokyo, Japan},
  year      = {1996},
  month     = mar,
  url       = {https://tufs.repo.nii.ac.jp/records/7640},
  note      = {In Japanese}
}

@article{liu-du-2008-tangut-processing,
  author  = {Liu, Changqing and Du, Jianlu},
  title   = {Web-Based Processing of the {Tangut} Script and
             {Tangut} Documents},
  journal = {Ningxia Social Sciences},
  number  = {5},
  pages   = {113--115},
  year    = {2008},
  issn    = {1002-0292},
  url     = {https://libproxy.qh.yitlink.com:8444/https/443/net/cnki/kns/yitlink/kcms2/article/abstract?v=lHKEv291v2jW7MUCwueKMJZat0U5-6bY4RaTRlgNK9I0wJlRYOSSy3ZaHIvI1HGRYzaC5nWGOyutp93Zi6Ebt0HnXgVh1wL1KyLUrL37F2sK_kid-b_lmBXE7bC9ENnt15RfKz15x4wP9cFWk9Nt9kfTgh4gZiYWqH9R0l95_fJ6ZE_eFIEbNA==&uniplatform=NZKPT&language=CHS},
  note    = {In Chinese}
}

@incollection{liu-2010-tangut-font-database,
  author    = {Liu, Changqing},
  title     = {Research on Constructing a Font Database for Historical
               {Tangut} Documents},
  booktitle = {Tangut Studies, Volume 6: Special Issue of the First
               International Forum on Tangut Studies, Part II},
  publisher = {Tangut Studies Research Institute, Ningxia University},
  pages     = {197--203},
  year      = {2010},
  url       = {https://www.sinoss.net/uploadfile/2012/0606/20120606103918739.pdf},
  note      = {In Chinese}
}

@techreport{west-etal-2014-tangut,
  author      = {West, Andrew and Everson, Michael and Han, Xiaomang and
                 Jia, Changye and Jing, Yongshi and Zaytsev, Viacheslav},
  title       = {Proposal to Encode the {Tangut} Script in the {UCS}},
  institution = {{ISO/IEC JTC 1/SC 2/WG 2}},
  number      = {N4522},
  year        = {2014},
  month       = jan,
  url         = {https://www.unicode.org/wg2/docs/n4522.pdf},
  note        = {Unicode document L2/14-023}
}

@article{liu-2012-tangut-document-recognition,
  author  = {Liu, Changqing},
  title   = {On {Tangut} Historical Documents Recognition},
  journal = {Physics Procedia},
  volume  = {33},
  pages   = {1212--1216},
  year    = {2012},
  doi     = {10.1016/j.phpro.2012.05.201},
  url     = {https://doi.org/10.1016/j.phpro.2012.05.201}
}

@inproceedings{zhang-han-2017-tangut-recognition,
  author    = {Zhang, Guangwei and Han, Xiaomang},
  title     = {Deep Learning Based {Tangut} Character Recognition},
  booktitle = {Proceedings of the 4th International Conference on
               Systems and Informatics},
  pages     = {437--441},
  publisher = {IEEE},
  year      = {2017},
  doi       = {10.1109/ICSAI.2017.8248332},
  url       = {https://doi.org/10.1109/ICSAI.2017.8248332}
}

@article{ma-etal-2022-tangut-recognition,
  author  = {Ma, Jinlin and Cao, Yunrui and Ma, Ziping and Wei, Lin and
             Hao, Chaohua},
  title   = {End-to-End {Tangut} Character Database Building and
             Recognition Method},
  journal = {IET Image Processing},
  volume  = {16},
  number  = {8},
  pages   = {2087--2100},
  year    = {2022},
  doi     = {10.1049/ipr2.12471},
  url     = {https://doi.org/10.1049/ipr2.12471}
}

@inproceedings{zheng-yu-2025-tangut-translation,
  author    = {Zheng, Yuxi and Yu, Jingsong},
  title     = {Incorporating Lexicon-Aligned Prompting in Large Language
               Models for {Tangut}--{Chinese} Translation},
  booktitle = {Proceedings of the Second Workshop on Ancient Language Processing},
  address   = {Albuquerque, New Mexico},
  publisher = {Association for Computational Linguistics},
  pages     = {127--136},
  year      = {2025},
  month     = may,
  doi       = {10.18653/v1/2025.alp-1.16},
  url       = {https://aclanthology.org/2025.alp-1.16/}
}

@article{zheng-etal-2025-tangut-ocr-mt,
  author  = {Zheng, Yuxi and Zhou, Ziming and Zhang, Yongwei and
             Sun, Bojun and Qiao, Wanxin and Hou, Junming and Yu, Jingsong},
  title   = {Research on {OCR} and {M}achine {T}ranslation for {T}angutScript under {L}ow-{R}esource {C}onditions},
  journal = {Digital Humanities},
  number  = {3},
  pages   = {113--135},
  year    = {2025},
  url     = {https://www.dhcn.cn/dhjournal/202503},
  note    = {In Chinese}
}

@article{shu-etal-2025-tangut-text-analysis,
  author  = {Shu, Xihong and Fan, Dandan and Wang, Yang},
  title   = {Textual Analysis of Excavated {Tangut} Documents in
             British and French Collections from a Digital Humanities
             Perspective},
  journal = {Journal of Dunhuang Studies},
  number  = {2},
  pages   = {115--130},
  year    = {2025},
  url     = {https://mp.weixin.qq.com/s/55lYYYk8pEWSvnpSB9H3ag},
  note    = {In Chinese}
}

@article{ sun-xixia,
author = { Sun, Sen},
title = {The {P}rocess of {D}igitizing the {T}angut {S}cript},
journal = {Corpus Linguistics},
number = {1},
pages = {130-139},
year = {2026},
url  = {https://kns.cnki.net/kcms2/article/abstract?v=BsQQ9aL8NZtLg5oZdKgjsyJpx-de5nUmdEPu1WiriVThpY2Q7UCde5MniJ9TaRtwhzTfb63mmtph6vD6t4gqdsXrKl7dzY55VdXLCf0jWskQVuxRxbM8uJKYNW1olNJ-3yCZ_Y8LKVSUPDoVtRlaBoc8-e2Wx6xvKpTB2HWIEhZ4hlikTT-HSQ==&uniplatform=NZKPT&language=CHS},
note = {In Chinese}
}

\clearpage

\appendix

\begin{CJK*}{UTF8}{gbsn}
\section{More Details About Our Corpus}
\label{sec:appendixA}

\subsection{Textual Sources and Composition}
\label{sec:appendixA1}

The annotated corpus is drawn from two Tangut works. The religious
portion comes from fascicle 68 of the Tangut translation of the
\textit{Mahāratnakūṭa Sūtra} (\textit{Da Bao Ji Jing},《大宝积经》), whereas the
secular portion comes from \textit{Leilin} (《类林》), an encyclopaedic collection of stories translated from
Chinese. Source-location identifiers preserve traceability to the
editions and catalogue records used by the annotators.

Due to the lack of an automatic sentence segmentation program, each segment is based on a single vertical column in the source
rubbing. Segment boundaries were adjusted
slightly where necessary to preserve lexical continuity across columns.
These units are therefore layout-based textual segments rather than
sentences defined by modern punctuation. Multiple specialists (especially Sen Sun) first analyzed the texts independently. Their boundary and label decisions were then compared, and disagreements were jointly reviewed to produce the consensus reference annotation. Formal IAA statistics were not recorded because this philological workflow was designed for adjudicated analysis rather than annotator measurement.

The two sources are substantially imbalanced. The religious text
accounts for approximately 8.5\% of the segments and 11.6\% of the
tokens, while \textit{Leilin} contributes the remainder. This imbalance
motivates our genre-specific evaluation.

The unlabeled corpus was extracted from the Tangut lines of digitized four-line translation materials based on manuscripts held in Russian collections. It contains 663 page-image records, covering 46 title-level units from seven works and approximately 318,000 Tangut characters. The largest source is \textit{Mahāratnakūṭa-sūtra} , whose 38 fascicle-level titles contribute 269,149 characters, or 84.6\% of the corpus. None of these records is from the annotated fascicle 68. Other Buddhist sources include \textit{《诸说禅源集都序》}, \textit{《注华严法界观门深入转》}, \textit{《中华传心地禅门师资承袭图》}, \textit{《修华严奥旨妄尽还源观》}, and \textit{《金师子章云间类解》}. The only clearly secular source is \textit{《六韬》}, represented by its first two volumes and accounting for approximately 1.3\% of the characters. Only the original Tangut text is used to compute distributional features and to train Char2Vec and TangutEncoder.

\subsection{Overlap and OOV-Coverage Audit}
\label{sec:overlap-audit}

We normalized both collections with the same Tangut-character filter and
split the unlabeled records at non-Tangut symbols. Five exact matches
remained, all one- or two-character fragments; no annotated segment of
three or more characters was duplicated. Table~\ref{tab:overlap-audit}
reports token-level $n$-gram overlap, defined as the percentage of
annotated $n$-gram occurrences whose character string appears anywhere
in the unlabeled collection.

\begin{table}[h]
\centering
\small
\setlength{\tabcolsep}{4pt}
\begin{tabular}{lrrr}
\toprule
Audit unit & All & Religious & Secular \\
\midrule
Exact segment ($\geq3$ chars) & 0 & 0 & 0 \\
5-gram overlap (\%)  & 0.37 & 2.66 & 0.02 \\
10-gram overlap (\%) & 0.01 & 0.08 & 0.00 \\
\bottomrule
\end{tabular}
\caption{Normalized overlap between annotated and unlabeled text.}
\label{tab:overlap-audit}
\end{table}

We also audit the resource exposure of labeled-training OOV tokens across
the five test folds. The external lexicon covers 53.0\%, and 34.8\%
occur as substrings in the unlabeled corpus; 12.6\% occur in both and
24.8\% in neither. These overlapping categories confirm that OOV-R is a
measure of generalization beyond supervised vocabulary, not exclusively
to forms unseen by all system components.

\subsection{Annotation Examples}

The following examples illustrate the corpus format. Spaces in the raw
line are removed, whereas the gold annotation marks word boundaries
with vertical bars and appends the linguistic label after a slash.

\begin{figure}[h]
    \centering
    \includegraphics[
        width=\columnwidth
    ]{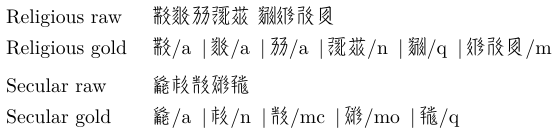}
    \caption{Annotation examples from the religious and secular
    portions of the corpus. Vertical bars indicate expert-annotated
    word boundaries.}
    \label{fig:corpus-examples}
\end{figure}

\subsection{POS and Morphosyntactic Labels}
\label{pos}

The expert annotation contains both conventional parts of speech and
more fine-grained morphosyntactic labels. After normalizing minor
orthographic variants, the current inventory contains 36 labels.
Frequent lexical categories include nouns (\texttt{n}), verbs
(\texttt{v}), adjectives (\texttt{a}), adverbs (\texttt{d}),
adpositions (\texttt{p}), conjunctions (\texttt{c}), pronouns
(\texttt{r}), numerals (\texttt{m}), quantifiers (\texttt{q}), and
particles or auxiliaries (\texttt{u}). The annotation additionally
preserves nominal subtypes and labels for grammatical functions,
aspect, direction, person, and number.

\begin{table}[h]
\centering
\small
\setlength{\tabcolsep}{3pt}
\begin{tabular}{p{0.27\columnwidth}p{0.65\columnwidth}}
\toprule
Group & Labels \\
\midrule
Core lexical categories
& \texttt{a, c, d, m, n, p, q, r, u, v} \\
\midrule
Fine-grained lexical labels
& \texttt{b, l, t, nb, nc, nh, nl, no, ns, mc, mo, rd, ri, rp} \\
\midrule
Morphosyntactic labels
& \texttt{Dir1., Dir2., Erg., Obj., Quot., Nom., Pfv.,
Fut., Loc., 1sg., 2sg., pl.} \\
\bottomrule
\end{tabular}
\caption{Normalized linguistic-label inventory.}
\label{tab:pos-inventory}
\end{table}

Because this inventory combines lexical classes with morphosyntactic
functions and remains under expert revision, the POS experiments in
this paper should be regarded as preliminary. We preserve the original
fine-grained distinctions rather than collapsing them into a newly
designed tagset.

\subsection{Independent Test Texts}
\label{sec:independent-test}

To evaluate cross-document generalization, we construct an independent
test set from five works absent from the supervised training corpus,
including three religious and two secular texts. The test set therefore
represents document-level unseen but within-domain evaluation rather than
transfer to previously unseen genres. As shown in
Table~\ref{tab:independent-test-data}, it contains 479 segments, 5,733
Tangut characters, and 4,081 word tokens.

\begin{table}[h]
\centering
\small
\setlength{\tabcolsep}{3.5pt}
\begin{tabular}{llrrrr}
\toprule
\textbf{Work} & \textbf{Genre} & \textbf{Seg.} &
\textbf{Tokens} & \textbf{Types} & \textbf{OOV} \\
\midrule
《华严经》
& Religious & 84  & 880   & 369 & 28.07\% \\
《妙法莲华经》
& Religious & 128 & 1380 & 457 & 20.22\% \\
《无量寿经》
& Religious & 90  & 1043 & 488 & 23.49\% \\
《德行集》
& Secular   & 71  & 406   & 203 & 21.92\% \\
《贤智集》
& Secular   & 106 & 372   & 257 & 37.63\% \\
\midrule
\textbf{Total} & --- & \textbf{479} & \textbf{4081} &
\textbf{1289} & \textbf{24.50\%} \\
\bottomrule
\end{tabular}
\caption{Composition of the independent cross-document test set.
OOV rates are computed over word tokens relative to the complete supervised training corpus. Word types in the total row are deduplicated across
the five works.}
\label{tab:independent-test-data}
\end{table}

Relative to the complete supervised training vocabulary used by the
evaluated models, 23.28\% of test tokens are OOV, indicating substantial
lexical novelty despite the low proportion of unseen characters.

We additionally audit possible textual overlap. The five source works
have no document-level overlap with the supervised corpus. Only three
test segments occur verbatim in the supervised data, all of which contain
one or two characters; no test segment of three or more characters is
duplicated. The longest shared substring contains nine characters, and
no ten-character test window occurs in the supervised training text.
Thus, there is no evidence of substantial leakage from the manually
segmented corpus.

The separation from unlabeled resources is less strict. No complete test
segment occurs verbatim in the MLM pretraining corpus, but eight segments
(1.67\%) contain a shared substring of at least eight characters and
three (0.63\%) contain one of at least ten characters. The longest case
is a 22-character overlap within a 24-character religious segment,
probably reflecting formulaic or parallel passages across Buddhist
works. Moreover, 73.86\% of multi-character supervised-OOV word types
occur as unsegmented strings in the unlabeled corpus, with substantially
higher coverage for religious than secular texts. The models may
therefore have encountered some character combinations, but not their
gold word boundaries. We accordingly describe this set as a
\emph{document-level unseen, within-domain test set}, rather than a
zero-exposure evaluation with respect to all external resources.

\section{Model and Training Parameters}
\label{sec:appendixB}

This appendix reports the architectural and optimization settings used in our experiments. Unless otherwise stated, all models use the same data partitions and BIES tag set. The base random seed is 42, and deterministic CuDNN execution is enabled within each training environment.

\subsection{Linear CRF}

The linear CRF is trained with L-BFGS. All CRF variants use the same optimization parameters and differ only in the external feature groups included in the input. The local feature template covers the current character, the two preceding and following characters, adjacent character bigrams, character-type indicators, and the single-character word indicator. Lexicon-lattice and corpus-statistical features are added as real-valued features.

\begin{table}[h]
\centering
\small
\setlength{\tabcolsep}{3pt}
\begin{tabular}{p{0.56\columnwidth}p{0.43\columnwidth}}
\toprule
Parameter & Value \\
\midrule
L1 coefficient ($c_1$) & $1.0$ \\
L2 coefficient ($c_2$) & $10^{-3}$ \\
Maximum L-BFGS iterations & 200 \\
Validation-based stopping & Not used \\
All possible transitions & Enabled \\
Character context window & $\pm 2$ characters \\
Inner folds for OOF reliability & 5 \\
Reliability smoothing ($\kappa$) & 5 \\
\bottomrule
\end{tabular}
\caption{Hyperparameters of the linear CRF. All reported CRF results use a fixed budget of 200 L-BFGS iterations.}
\label{tab:crf-parameters}
\end{table}

\subsection{BiLSTM--CRF}

Before introducing external knowledge, we conducted a preliminary hyperparameter search for the supervised BiLSTM--CRF. We varied the number of LSTM layers, character-embedding dimension, BiLSTM hidden size, batch size, dropout rate, and learning rate. This search used a fixed 80/10/10 split created before the outer cross-validation and was therefore not nested within each outer training fold. The selected configuration was subsequently fixed across all folds. We accordingly treat the BiLSTM as a supporting architectural comparison rather than use it to substantiate the main model-selection claim; the CRF and TangutEncoder comparisons do not use this search.

Table~\ref{tab:bilstm-search} reports the complete search results. A two-layer BiLSTM with 100-dimensional character embeddings and a total hidden size of 64 achieved the highest observed segmentation F$_1$. Among the tested batch sizes, 512 produced the best F$_1$, although a batch size of 32 yielded a similar result. We therefore selected the former for its higher training throughput. 

\begin{table}[h]
\centering
\scriptsize
\begin{tabular}{rrrrrrccc}
\toprule
Layers & Emb. & Hidden & Batch & Dropout & LR
& F$_1$ \\
\midrule
2 & 100 & 256 & 512  & 0.3 & $5\times10^{-4}$ & 0.8636\\
2 & 100 & 128 & 512  & 0.3 & $5\times10^{-4}$ & 0.8595\\
\textbf{2} & \textbf{100} & \textbf{64} & \textbf{512}
& \textbf{0.3} & $\mathbf{5\times10^{-4}}$
& \textbf{0.8905} \\
2 & 100 & 32  & 512  & 0.3 & $5\times10^{-4}$ & 0.8873 \\
2 & 100 & 16  & 512  & 0.3 & $5\times10^{-4}$ & 0.8423 \\
\midrule
1 & 100 & 64  & 512  & 0.3 & $5\times10^{-4}$ & 0.8638 \\
1 & 100 & 128 & 512  & 0.3 & $5\times10^{-4}$ & 0.8569 \\
3 & 100 & 64  & 512  & 0.3 & $5\times10^{-4}$ & 0.8659 \\
\midrule
2 & 100 & 64 & 1024 & 0.3 & $5\times10^{-4}$ & 0.8829 \\
2 & 100 & 64 & 256  & 0.3 & $5\times10^{-4}$ & 0.8822 \\
2 & 100 & 64 & 128  & 0.3 & $5\times10^{-4}$ & 0.8861 \\
2 & 100 & 64 & 64   & 0.3 & $5\times10^{-4}$ & 0.8870 \\
2 & 100 & 64 & 32   & 0.3 & $5\times10^{-4}$ & 0.8900  \\
2 & 100 & 64 & 16   & 0.3 & $5\times10^{-4}$ & 0.8854 \\
2 & 100 & 64 & 32   & 0.3 & $1\times10^{-4}$ & 0.8847\\
\midrule
2 & 32  & 64 & 512 & 0.3 & $5\times10^{-4}$ & 0.8382 \\
2 & 64  & 64 & 512 & 0.3 & $5\times10^{-4}$ & 0.8496 \\
2 & 128 & 64 & 512 & 0.3 & $5\times10^{-4}$ & 0.8814 \\
2 & 256 & 64 & 512 & 0.3 & $5\times10^{-4}$ & 0.8755 \\
\midrule
2 & 100 & 64 & 512 & 0.1 & $5\times10^{-4}$ & 0.8824 \\
2 & 100 & 64 & 512 & 0.5 & $5\times10^{-4}$ & 0.8870 \\
\bottomrule
\end{tabular}
\caption{Preliminary hyperparameter search for the supervised
BiLSTM--CRF. Emb.\ denotes the character-embedding dimension, and
Hidden denotes the concatenated output dimension of the two directions.
The best segmentation F$_1$ is highlighted.}
\label{tab:bilstm-search}
\end{table}

Based on this search, the final BiLSTM--CRF uses randomly initialized and trainable 100-dimensional character embeddings, followed by a two-layer bidirectional LSTM. Each direction has 32 hidden units, producing a 64-dimensional contextual representation. A linear layer maps this representation to four BIES emission scores, which are globally decoded by a CRF.

When external knowledge is enabled, the corresponding feature vectors are concatenated directly with the character embeddings before the BiLSTM. The principal lexicon-enhanced model uses the 20-dimensional reliability-calibrated dictionary representation. A dictionary dropout rate of 0.2 is applied during training to reduce excessive reliance on lexical matches.

\begin{table}[h]
\centering
\small
\setlength{\tabcolsep}{3pt}
\begin{tabular}{p{0.46\columnwidth}p{0.43\columnwidth}}
\toprule
Parameter & Selected value \\
\midrule
Character embedding size & 100 \\
BiLSTM layers & 2 \\
Hidden size & 32 per direction \\
BiLSTM output size & 64 \\
Emission size & 4 \\
Model dropout & 0.3 \\
Dictionary dropout & 0.2 \\
Parameter initialization & Xavier uniform \\
Optimizer & Adam \\
Learning rate & $5\times10^{-4}$ \\
Batch size & 512 sentences \\
Maximum epochs & 10,000 \\
Early-stopping patience & 3 epochs \\
Selection criterion & Development NLL \\
Gradient clipping & 5.0 \\
\bottomrule
\end{tabular}
\caption{Selected architectural and training parameters of the
BiLSTM--CRF. The maximum number of epochs is only an upper bound;
the checkpoint with the lowest development loss is restored after
early stopping.}
\label{tab:bilstm-parameters}
\end{table}

\subsection{Transformer--CRF Variants}

Transformer-Random, Transformer-Char2Vec, and TangutEncoder (TEnc) share the same Transformer and CRF architecture. They differ only in parameter initialization: Transformer-Random initializes the complete encoder randomly; Transformer-Char2Vec initializes its character-embedding matrix with static Skip-gram vectors; and TangutEncoder loads all encoder parameters from masked-language-model pretraining.

The encoder contains three pre-normalized Transformer layers with a hidden size of 192, four attention heads, and a 768-dimensional feed-forward layer. Learned position embeddings support sequences of up to 128 characters. The encoder output is mapped to four BIES emission scores and decoded by a CRF. With the current vocabulary, the encoder contains approximately 2.5 million parameters, although the exact number varies with vocabulary size.

\begin{table}[h]
\centering
\small
\setlength{\tabcolsep}{3pt}
\begin{tabular}{p{0.46\columnwidth}p{0.43\columnwidth}}
\toprule
Parameter & Value \\
\midrule
Character embedding size & 192 \\
Maximum sequence length & 128 \\
Transformer layers & 3 \\
Attention heads & 4 \\
Dimension per head & 48 \\
Feed-forward size & 768 \\
Activation & GELU \\
Normalization & Pre-LN \\
Encoder dropout & 0.15 \\
Task-head dropout & 0.20 \\
Position embeddings & Learned \\
Dictionary projection & $20\rightarrow32$ \\
Distributional projection & $8\rightarrow16$ \\
Output layer & Linear + CRF \\
\bottomrule
\end{tabular}
\caption{Shared architecture of the Transformer--CRF variants. The two external-feature projections are included only in the corresponding fusion models.}
\label{tab:transformer-architecture}
\end{table}

\paragraph{Char2Vec initialization.}
Char2Vec is trained on character sequences extracted from the unlabeled corpus using Skip-gram with negative sampling. Its vectors are rescaled to match the standard deviation of the randomly initialized Transformer embeddings before being copied into the embedding matrix. Characters not covered by Char2Vec remain randomly initialized.

\begin{table}[h]
\centering
\small
\setlength{\tabcolsep}{3pt}
\begin{tabular}{p{0.46\columnwidth}p{0.43\columnwidth}}
\toprule
Char2Vec parameter & Value \\
\midrule
Training objective & Skip-gram \\
Vector size & 192 \\
Context window & 5 \\
Negative samples & 10 \\
Minimum frequency & 1 \\
Training epochs & 30 \\
Workers & 1 \\
Random seed & 42 \\
Minimum sequence length & 2 characters \\
\bottomrule
\end{tabular}
\caption{Training parameters of the static Char2Vec initialization.}
\label{tab:char2vec-parameters}
\end{table}

\paragraph{TangutEncoder pretraining.}
TangutEncoder is pretrained with character-level masked language modeling. Approximately 15\% of character positions are selected, with equal probability of single-character masking and contiguous span masking of two to four characters. Selected positions follow the standard 80/10/10 replacement strategy. Sequences longer than 128 characters are divided into consecutive non-overlapping chunks; all labeled instances are shorter than this limit. The vocabulary is constructed from the lexicon and unlabeled corpus.

\begin{table}[h]
\centering
\small
\setlength{\tabcolsep}{3pt}
\begin{tabular}{p{0.46\columnwidth}p{0.43\columnwidth}}
\toprule
MLM parameter & Value \\
\midrule
Masking ratio & 0.15 \\
Span-masking probability & 0.50 \\
Span length & 2--4 characters \\
Replacement strategy & 80/10/10 \\
Optimizer & AdamW \\
Learning rate & $3\times10^{-4}$ \\
Weight decay & 0.01 \\
Batch size & 32 sequences \\
Maximum steps & 5,000 \\
Warm-up steps & 500 \\
Learning-rate schedule & Warm-up then constant \\
Evaluation interval & 200 steps \\
Early-stopping patience & 5 evaluations \\
Gradient clipping & 1.0 \\
Validation split & 5 held-out UUIDs \\
\bottomrule
\end{tabular}
\caption{Masked-language-model pretraining parameters of TangutEncoder. Pretraining completed the full 5,000 steps, and the checkpoint with the lowest validation loss was retained.}
\label{tab:mlm-parameters}
\end{table}

\paragraph{Downstream fine-tuning.}
All three Transformer variants use the same two-stage downstream procedure. The encoder is first frozen for three epochs while the randomly initialized task head and CRF are trained, and is then jointly fine-tuned. This shared schedule controls the pipeline but is primarily designed for pretrained encoders; the random and Char2Vec variants are therefore initialization controls rather than independently optimized scratch upper bounds.

\begin{table}[h]
\centering
\small
\setlength{\tabcolsep}{3pt}
\begin{tabular}{p{0.5\columnwidth}p{0.43\columnwidth}}
\toprule
Fine-tuning parameter & Value \\
\midrule
Frozen-encoder epochs & 3 \\
Optimizer & AdamW \\
Encoder learning rate & $5\times10^{-5}$ \\
Task-head learning rate & $5\times10^{-4}$ \\
Weight decay & 0.01 \\
Batch size & 32 sentences \\
Maximum fine-tuning epochs & 10,000 \\
LR-reduction patience & 5 epochs \\
LR-reduction factor & 0.3 \\
Early-stopping patience & 10 epochs \\
Selection criterion & Development NLL \\
Gradient clipping & 1.0 \\
\bottomrule
\end{tabular}
\caption{Downstream training parameters shared by Transformer--Random, Transformer--Char2Vec and TangutEncoder.}
\label{tab:transformer-finetuning}
\end{table}

\section{More Experimental Data}
\label{sec:appendixC}

\subsection{Comparison of Character Association Measures}
\label{sec:association-measures}

The explicit distributional representation includes an association score for each adjacent character pair. Let $f(a,b)$ denote the frequency of bigram $(a,b)$, $f(a)$ and $f(b)$ the corresponding unigram frequencies, and $N$ the total number of adjacent character pairs in the unlabeled corpus. We compare three association measures.

Discounted pointwise mutual information reduces the disproportionate scores that standard PMI assigns to very rare bigrams:
\begin{equation*}
\operatorname{dPMI}(a,b)
=
\max\left(
0,\,
\log
\frac{
\left(f(a,b)-\delta\right)N
}{
f(a)f(b)
}
\right).
\end{equation*}

The subtraction of $\delta=0.5$ discounts low-frequency observations, while truncation at zero retains only positive association. This measure emphasizes pairs occurring more frequently than expected under character independence.

The Dice coefficient measures the normalized overlap between the occurrence distributions of two characters:
\begin{equation*}
\operatorname{Dice}(a,b)
=
\frac{2f(a,b)}
{f(a)+f(b)}.
\end{equation*}

Dice is bounded between zero and one and is less susceptible than PMI to extremely high values caused by rare events. However, it does not explicitly compare the observed bigram frequency with an independence-based expectation.

Finally, the $t$-score measures the standardized difference between observed and expected co-occurrence:
\begin{equation*}
t(a,b)
=
\frac{
f(a,b)-\dfrac{f(a)f(b)}{N}
}{
\sqrt{f(a,b)}
}.
\end{equation*}

Unlike positive dPMI, the $t$-score can preserve negative evidence when a pair occurs less frequently than expected. It also tends to favor associations supported by higher absolute frequencies.

Table~\ref{tab:association-comparison} compares these measures under the same CRF, lexicon features, and bigram-frequency features. For each measure, we report results both before and after adding neighbor entropy.

\begin{table*}[h]
\centering
\footnotesize
\setlength{\tabcolsep}{4pt}
\begin{tabular}{lccccc}
\toprule
Association features & P & R & F$_1$ & OOV-R & IV-R \\
\midrule
Freq + dPMI
& $0.906{\pm}0.003$ & $0.905{\pm}0.003$
& $0.905{\pm}0.003$ & $0.486{\pm}0.009$
& $0.945{\pm}0.001$ \\
Freq + dPMI + Ent
& $0.907{\pm}0.004$ & $0.906{\pm}0.004$
& $0.907{\pm}0.004$ & $0.491{\pm}0.016$
& $0.947{\pm}0.002$ \\
\midrule
Freq + Dice
& $0.905{\pm}0.003$ & $0.905{\pm}0.004$
& $0.905{\pm}0.003$ & $0.485{\pm}0.010$
& $0.945{\pm}0.002$ \\
Freq + Dice + Ent
& $0.907{\pm}0.004$ & $0.906{\pm}0.004$
& $0.906{\pm}0.004$ & $0.489{\pm}0.016$
& $0.946{\pm}0.002$ \\
\midrule
Freq + $t$-score
& $0.905{\pm}0.004$ & $0.904{\pm}0.004$
& $0.904{\pm}0.004$ & $0.483{\pm}0.014$
& $0.944{\pm}0.003$ \\
Freq + $t$-score + Ent
& $0.907{\pm}0.003$ & $0.906{\pm}0.004$
& $0.907{\pm}0.004$ & $0.490{\pm}0.013$
& $0.946{\pm}0.002$ \\
\bottomrule
\end{tabular}
\caption{Comparison of association measures under five-fold cross-validation.}
\label{tab:association-comparison}
\end{table*}

Without entropy, dPMI performs slightly better than Dice and $t$-score, particularly on OOV words. After adding entropy, however, all three measures converge to nearly identical performance. The results therefore do not establish a statistically meaningful winner.

The consistent improvement from neighbor entropy also indicates that association strength and contextual diversity provide complementary evidence. Association measures estimate whether two adjacent characters tend to form a cohesive unit, whereas entropy indicates whether either character occurs freely with many different neighbors, which is more directly related to potential word boundaries. We use dPMI in the main experiments because its frequency discount is appropriate for the many low-count bigrams in the Tangut corpus and because it provides a favorable balance between overall performance and OOV recall.

\subsection{Additional Model Variants}
\label{sec:additional-variants}

\paragraph{Dictionary sources.}

The main text reports maximum matching with the training-corpus vocabulary. We additionally compare three sources: the training vocabulary (\textsc{Dict-corpus}), the external Tangut dictionary alone (\textsc{Dict-dictionary}), and their union (\textsc{Dict-all}). All three settings use bidirectional maximum matching. (Table~\ref{tab:dictionary-source})

\begin{table}[h]
\centering
\small
\setlength{\tabcolsep}{3pt}
\begin{tabular}{lccccc}
\toprule
Model & P & R & F$_1$ & OOV-R & IV-R \\
\midrule
Dict-corpus
& .845 & .890 & .867 & .219 & .954 \\
Dict-dictionary
& .787 & .728 & .756 & .666 & .733 \\
Dict-all
& .821 & .740 & .779 & .665 & .748 \\
\bottomrule
\end{tabular}
\caption{Bidirectional maximum matching with different dictionary sources. Values are five-fold means.}
\label{tab:dictionary-source}
\end{table}

\begin{table*}[h]
\centering
\footnotesize
\setlength{\tabcolsep}{3.5pt}
\begin{tabular}{lcccccc}
\toprule
Model & External features & P & R & F$_1$ & OOV-R & IV-R \\
\midrule
BiLSTM
& 0
& $0.862{\pm}0.008$ & $0.873{\pm}0.008$
& $0.868{\pm}0.008$ & $0.472{\pm}0.014$
& $0.912{\pm}0.006$ \\
BiLSTM+BIE
& 11
& $0.885{\pm}0.005$ & $0.894{\pm}0.004$
& $0.889{\pm}0.004$ & $0.546{\pm}0.011$
& $0.927{\pm}0.004$ \\
BiLSTM+BIE+R
& 14
& $0.891{\pm}0.003$ & $0.898{\pm}0.004$
& $0.894{\pm}0.004$ & $0.546{\pm}0.009$
& $0.931{\pm}0.003$ \\
BiLSTM+$\mathrm{Dict}_{\mathrm{all}}$
& 17
& $0.893{\pm}0.003$ & $0.900{\pm}0.003$
& $0.897{\pm}0.003$ & $0.551{\pm}0.011$
& $0.933{\pm}0.003$ \\
\midrule
\quad + $\mathrm{Dict}_{\mathrm{all}}$ + IL
& 28
& $0.896{\pm}0.003$ & $0.900{\pm}0.002$
& $0.898{\pm}0.002$ & $0.557{\pm}0.009$
& $0.933{\pm}0.002$ \\
\quad + $\mathrm{Dict}_{\mathrm{all}}$ + IL + DD
& 30
& $0.897{\pm}0.004$ & $0.904{\pm}0.003$
& $0.900{\pm}0.003$ & $0.554{\pm}0.006$
& $0.938{\pm}0.002$ \\
\quad + $\mathrm{Dict}_{\mathrm{all}}$ + IL + DD + Freq
& $30+2$
& $0.902{\pm}0.004$ & $0.909{\pm}0.005$
& $0.906{\pm}0.004$ & $0.560{\pm}0.008$
& $0.943{\pm}0.005$ \\
\quad + $\mathrm{Dict}_{\mathrm{all}}$ + IL + DD + Freq + Coo
& $30+4$
& $0.902{\pm}0.004$ & $0.909{\pm}0.004$
& $0.905{\pm}0.003$ & $0.563{\pm}0.016$
& $0.941{\pm}0.004$ \\
\quad + $\mathrm{Dict}_{\mathrm{all}}$ + IL + DD + $\mathrm{Dist}_{\mathrm{all}}$
& $30+8$
& $0.902{\pm}0.003$ & $0.909{\pm}0.004$
& $0.905{\pm}0.003$ & $0.562{\pm}0.014$
& $0.942{\pm}0.004$ \\
\bottomrule
\end{tabular}
\caption{BiLSTM--CRF variants with progressively richer external features.}
\label{tab:bilstm-feature-variants}
\end{table*}

\begin{table*}[h]
\centering
\footnotesize
\setlength{\tabcolsep}{4pt}
\begin{tabular}{lcccc}
\toprule
& \multicolumn{2}{c}{Secular} & \multicolumn{2}{c}{Religious} \\
Model & F$_1$ & OOV-R & F$_1$ & OOV-R \\
\midrule
BiLSTM
& $0.875{\pm}0.007$ & $0.498{\pm}0.016$
& $0.810{\pm}0.021$ & $0.244{\pm}0.028$ \\
BiLSTM+BIE
& $0.897{\pm}0.005$ & $0.573{\pm}0.015$
& $0.830{\pm}0.018$ & $0.305{\pm}0.043$ \\
BiLSTM+BIE+R
& $0.903{\pm}0.004$ & $0.574{\pm}0.014$
& $0.829{\pm}0.013$ & $0.286{\pm}0.053$ \\
BiLSTM+$\mathrm{Dict}_{\mathrm{all}}$
& $0.905{\pm}0.003$ & $0.579{\pm}0.014$
& $0.835{\pm}0.016$ & $0.298{\pm}0.037$ \\
\quad + $\mathrm{Dict}_{\mathrm{all}}$ + IL
& $0.905{\pm}0.002$ & $0.586{\pm}0.011$
& $0.842{\pm}0.018$ & $0.300{\pm}0.043$ \\
\quad + $\mathrm{Dict}_{\mathrm{all}}$ + IL + DD
& $0.907{\pm}0.001$ & $0.586{\pm}0.008$
& $0.850{\pm}0.021$ & $0.270{\pm}0.041$ \\
\quad + $\mathrm{Dict}_{\mathrm{all}}$ + IL + DD + Freq
& $0.911{\pm}0.003$ & $0.588{\pm}0.012$
& $0.862{\pm}0.019$ & $0.316{\pm}0.056$ \\
\quad + $\mathrm{Dict}_{\mathrm{all}}$ + IL + DD + Freq + Coo
& $0.911{\pm}0.002$ & $0.591{\pm}0.023$
& $0.859{\pm}0.026$ & $0.322{\pm}0.076$ \\
\quad + $\mathrm{Dict}_{\mathrm{all}}$ + IL + DD + $\mathrm{Dist}_{\mathrm{all}}$
& $0.911{\pm}0.001$ & $0.589{\pm}0.020$
& $0.860{\pm}0.021$ & $0.326{\pm}0.051$ \\
\bottomrule
\end{tabular}
\caption{Genre-level results for the BiLSTM--CRF feature variants.}
\label{tab:bilstm-genre-variants}
\end{table*}

The training vocabulary provides strong IV recall but has low recall beyond the labeled vocabulary. Conversely, the external dictionary substantially increases OOV recall, but its lower precision and IV recall indicate frequent disagreement with corpus boundaries. Thus, greater lexical coverage does not automatically produce better segmentation, motivating the aggregated lattice representation and reliability calibration.

\paragraph{Alternative CRF lattice features.}

We also examined candidate-count and maximum-candidate-length features in an earlier lattice design. Adding candidate counts to the BIE lattice produced an F$_1$ of 0.898, compared with 0.897 for BIE alone, while maximum-length features produced an F$_1$ of 0.897. 

\paragraph{BiLSTM--CRF feature variants.}

We additionally examined two feature blocks that were used only in the
BiLSTM--CRF experiments and were therefore excluded from the main
framework.

\textit{Internal lattice }(\textbf{IL}).
The external dictionary does not contain all corpus-specific words.
We therefore construct an internal lexicon from multi-character words
observed in the outer training split but absent from the external
dictionary:
\[
\mathcal{D}_{\mathrm{int}}
=
\left\{
w \in \mathcal{V}_{\mathrm{train}}:
|w|\geq 2,\,
w\notin\mathcal{D}_{\mathrm{ext}}
\right\}.
\]
All matches from $\mathcal{D}_{\mathrm{int}}$ are encoded using the same
11-dimensional BIE-by-length scheme as the external lattice
($B2$--$B5+$, $I3$--$I5+$, and $E2$--$E5+$).
Unlike the external dictionary features, these dimensions contain no
reliability scores, because their entries are derived directly from the
annotated corpus. To prevent gold boundaries from being revealed during
training, the internal lexicon features are generated by inner
five-fold out-of-fold estimation; development and test features are
constructed from the complete outer training split. Combining this
11-dimensional block with $\mathrm{Dict}_{\mathrm{all}}$ produces a
28-dimensional lexical representation.

\textit{Domain distribution }(\textbf{DD}).
To represent the domain preference of a candidate word, we count its
occurrences in the religious and secular portions of the outer training
data and define
\[
\textbf{q}(w)
=
\left[
\frac{n_{\mathrm{rel}}(w)}
     {n_{\mathrm{rel}}(w)+n_{\mathrm{sec}}(w)},
\frac{n_{\mathrm{sec}}(w)}
     {n_{\mathrm{rel}}(w)+n_{\mathrm{sec}}(w)}
\right].
\]
At each character position, the vectors of all external- and
internal-lexicon candidates covering that position are averaged,
yielding two additional dimensions. If no matched candidate has an
estimated distribution, the implementation falls back to the known
document genre, represented as $[1,0]$ for religious texts and $[0,1]$
for secular texts. The word distributions and training features are
again estimated out of fold. Adding this block to the 28-dimensional
lexical representation gives 30 dimensions in total.

The internal lattice is intended to recover corpus-specific IV words
missing from the traditional dictionary, whereas the domain
distribution indicates whether a candidate is associated with
religious or secular usage. The latter assumes that document genre is
available at inference time; it is therefore treated as an auxiliary
analysis rather than part of the main genre-independent model.

Table~\ref{tab:bilstm-feature-variants} reports the complete progression of external features in the BiLSTM--CRF. The feature dimension excludes the 100-dimensional character embedding. The ablations show a clear hierarchy of feature contributions. The BIE
lattice provides the largest lexical gain, followed by reliability
estimation and priors for unseen entries. The internal lattice and domain
features offer smaller complementary improvements by recovering
corpus-specific words and modeling genre preferences. Among the
unlabeled-text features, bigram frequency contributes most, whereas
association and entropy provide little additional benefit, possibly
because the BiLSTM already captures part of this local information.

As shown in Table~\ref{tab:bilstm-genre-variants}, the improvements are
more stable on secular texts because they dominate the training data.
Nevertheless, the internal lattice and domain features also improve
religious-text performance, including OOV recall. Since the
domain features assume that document genre is known at inference time,
they are retained only as an auxiliary variant.

\paragraph{Additional Transformer controls.}

Injecting the lexicon lattice into a randomly initialized Transformer improves both F$_1$ and OOV recall over the corresponding baseline. However, it remains below the MLM-pretrained encoder, showing that dictionary knowledge cannot substitute for contextual pretraining. Conversely, combining MLM pretraining with the lexicon representation produces a substantially stronger model than either source alone.

\subsection{Lexicon-aware Continued Pretraining}

We additionally continued MLM pretraining on TangutEncoder with a weakly supervised
dictionary-span ranking objective. Dictionary entries of two to four
characters from the structured lexicon and the revised \textit{Tongyin}(《同音》)
were matched against the unlabeled corpus and treated as positive
candidate spans. For each positive span, five non-dictionary spans were
sampled from the same sequence as negatives.

For a span $(i,j)$, the scoring head concatenates boundary-sensitive
states, mean and max pooling, and a length embedding:
\[
\begin{aligned}
\textbf{v}_{i,j}=[&
\textbf{h}_i;\textbf{h}_{j-1};
\operatorname{Mean}(\textbf{h}_{i:j});\\
&\operatorname{Max}(\textbf{h}_{i:j});
\textbf{e}_{j-i+1}],
\end{aligned}
\]
and assigns the span a scalar score
$s(i,j)=\operatorname{MLP}(\textbf{v}_{i,j})$.
The pairwise ranking loss is
\[
\mathcal{L}_{\mathrm{word}}
=
-\frac{1}{|\mathcal{P}|K}
\sum_{p\in\mathcal{P}}
\sum_{n\in\mathcal{N}(p)}
\log\sigma\!\left(s(p)-s(n)\right).
\]
It is combined with MLM as
\[
\mathcal{L}
=
\mathcal{L}_{\mathrm{MLM}}
+
\lambda_{\mathrm{word}}\mathcal{L}_{\mathrm{word}},
\]
where $\lambda_{\mathrm{word}}=0.3$. Training continued for at most
3,000 steps, with the word-loss weight linearly warmed up during the
first 500 steps. The span head was discarded after pretraining, and
only the updated encoder was transferred to the segmentation model.

\begin{table}[h]
\centering
\small
\setlength{\tabcolsep}{4pt}
\begin{tabular}{lccc}
\toprule
Pretraining & F$_1$ & OOV-R & IV-R \\
\midrule
MLM
& $0.911$
& $0.608$
& $0.946$ \\
MLM + WordRank
& $0.912$
& $0.617$
& $0.946$ \\
\bottomrule
\end{tabular}
\caption{Downstream results of lexicon-aware continued pretraining.
Both encoders are evaluated with the same dictionary and distributional
features.}
\label{tab:word-ranking}
\end{table}

The auxiliary objective yields only marginal numerical gains in F$_1$
and OOV recall, within cross-fold variation. Because it adds weak-supervision
assumptions without a consistent downstream gain, we report it only as
a preliminary appendix experiment.

\subsection{Full Results on Unseen Documents}
\label{sec:appendixC4}

Table~\ref{tab:unseen-full-results} reports the complete results on the
independent test set introduced in Appendix~\ref{sec:independent-test}.

\begin{table*}[t]
\centering
\scriptsize
\setlength{\tabcolsep}{4.2pt}
\renewcommand{\arraystretch}{0.94}
\begin{tabular}{lrrrrrr}
\toprule
\textbf{Model} &
\textbf{P} & \textbf{R} & \textbf{F$_1$} &
\textbf{POS Acc.} & \textbf{POS $N$} \\
\midrule

\multicolumn{6}{l}{\textit{Overall (479 segments)}} \\
CRF
& 0.6410 & 0.7381 & 0.6861 & -- & -- \\
CRF + $\mathrm{Dict}_{\mathrm{all}}$
& 0.6850 & 0.7711 & 0.7255 & -- & -- \\
CRF + $\mathrm{Dict}_{\mathrm{all}}$
    + $\mathrm{Dist}_{\mathrm{all}}$
& 0.7114 & 0.7863 & 0.7470 & -- & -- \\
Transformer-Random
& 0.5915 & 0.7008 & 0.6415 & -- & -- \\
Transformer-Char2Vec
& 0.6015 & 0.7121 & 0.6522 & -- & -- \\
TangutEncoder
& 0.6806 & 0.7692 & 0.7222 & -- & -- \\
TangutEncoder + $\mathrm{Dict}_{\mathrm{all}}$
    + $\mathrm{Dist}_{\mathrm{all}}$
& 0.7079 & \textbf{0.8069} & 0.7542 & -- & -- \\
POS-tagging Model(Joint CRF) 
& \textbf{0.7266} & 0.7873 & \textbf{0.7557}
& \textbf{0.8245} & 3,191 \\
\midrule

\multicolumn{6}{l}{\textit{Religious texts (302 segments)}} \\
CRF
& 0.6389 & 0.7445 & 0.6876 & -- & -- \\
CRF + $\mathrm{Dict}_{\mathrm{all}}$
& 0.6888 & 0.7835 & 0.7331 & -- & -- \\
CRF + $\mathrm{Dict}_{\mathrm{all}}$
    + $\mathrm{Dist}_{\mathrm{all}}$
& \textbf{0.7207} & 0.7999 & 0.7582 & -- & -- \\
Transformer-Random
& 0.5878 & 0.7015 & 0.6396 & -- & -- \\
Transformer-Char2Vec
& 0.5996 & 0.7115 & 0.6508 & -- & -- \\
TangutEncoder
& 0.6871 & 0.7793 & 0.7303 & -- & -- \\
TangutEncoder+ $\mathrm{Dict}_{\mathrm{all}}$
    + $\mathrm{Dist}_{\mathrm{all}}$
& 0.7149 & \textbf{0.8123} & \textbf{0.7605} & -- & -- \\
POS-tagging Model(Joint CRF)
& \textbf{0.7266} & 0.7950 & 0.7593
& \textbf{0.8447} & 2,615 \\
\midrule

\multicolumn{6}{l}{\textit{Secular texts (177 segments)}} \\
CRF
& 0.6506 & 0.7108 & 0.6794 & -- & -- \\
CRF + $\mathrm{Dict}_{\mathrm{all}}$
& 0.6679 & 0.7185 & 0.6923 & -- & -- \\
CRF + $\mathrm{Dict}_{\mathrm{all}}$
    + $\mathrm{Dist}_{\mathrm{all}}$
& 0.6710 & 0.7288 & 0.6987 & -- & -- \\
Transformer-Random
& 0.6081 & 0.6979 & 0.6499 & -- & -- \\
Transformer-Char2Vec
& 0.6096 & 0.7147 & 0.6580 & -- & -- \\
TangutEncoder
& 0.6524 & 0.7262 & 0.6873 & -- & -- \\
TangutEncoder + $\mathrm{Dict}_{\mathrm{all}}$
    + $\mathrm{Dist}_{\mathrm{all}}$
& 0.6785 & \textbf{0.7841} & 0.7275 & -- & -- \\
POS-tagging Model(Joint CRF)
& \textbf{0.7265} & 0.7545 & \textbf{0.7402}
& \textbf{0.7326} & 576 \\
\bottomrule
\end{tabular}
\caption{Full segmentation and POS-tagging results on five unseen works. POS accuracy is calculated only over words whose predicted boundaries exactly match the gold segmentation; POS $N$ denotes the number of such words.}
\label{tab:unseen-full-results}
\end{table*}

The independent evaluation confirms the contribution of both external
knowledge and contextual pretraining. For the linear CRF, adding the
lexicon representation and distributional statistics progressively
raises F$_1$ from 0.6861 to 0.7470. Char2Vec provides only a small gain
over random Transformer initialization, whereas MLM pretraining produces
a substantially larger improvement. Combining MLM with lexical and
distributional features further raises F$_1$ to 0.7542.

The joint CRF obtains the highest overall F$_1$ of 0.7557, only
marginally above the complete TangutEncoder. 
Genre-specific results show that the complete TangutEncoder performs
best on religious texts, whereas linear CRF is stronger on secular
texts. POS-tagging Model's conditional POS accuracy reaches 0.8245 overall, but is
considerably lower on secular texts, indicating that POS transfer
remains sensitive to genre and annotation coverage.

\clearpage

\clearpage
\onecolumn

\section{Tangut Digital Intelligence Platform}
\label{sec:appendixD}

\noindent
The platform is available at
\url{http://lmjx.ucass.edu.cn/xixia/}.

\begin{figure}[h]
    \centering

    \includegraphics[width=0.78\textwidth]{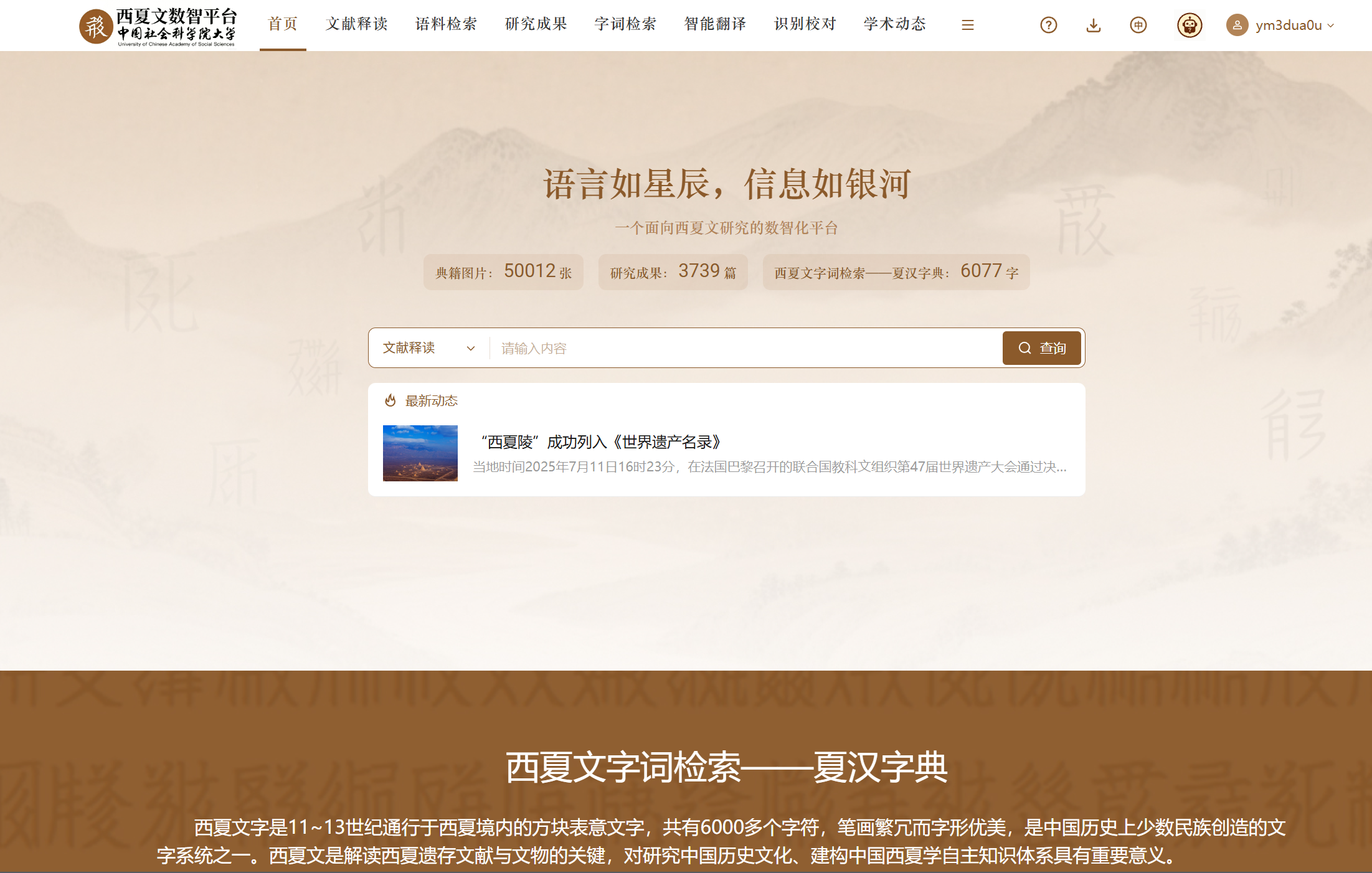}

    \vspace{2mm}
    {\small\textbf{(a)} Homepage of the Tangut Digital Intelligence Platform.}

    \vspace{4mm}

    \includegraphics[width=0.78\textwidth]{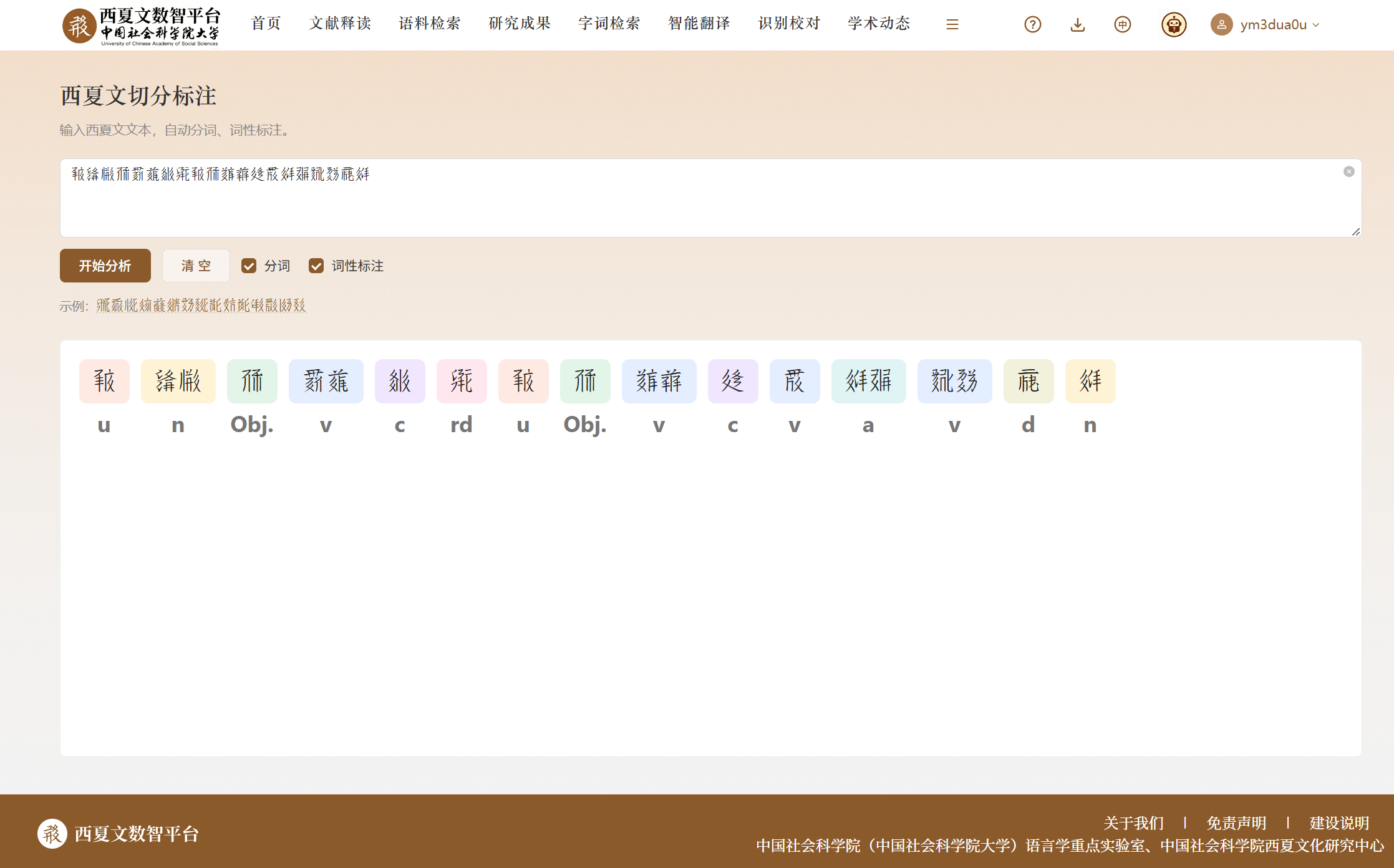}

    \vspace{2mm}
    {\small\textbf{(b)} Online Tangut word segmentation and POS-tagging interface.}

    \caption{The Tangut Digital Intelligence Platform.}
    \label{fig:tangut-platform}
\end{figure}
\end{CJK*}
\end{document}